\documentclass{article} 

\usepackage{iclr2026_conference}
\usepackage{times}

\usepackage{newtxtt} 
\usepackage[scaled=1]{biolinum} 
\usepackage{graphicx}
\usepackage{booktabs} 
\usepackage[font={footnotesize}]{caption}
\usepackage{subcaption} 
\usepackage{float} 
\usepackage[utf8]{inputenc} 
\usepackage[T1]{fontenc}    
\usepackage[colorlinks=true, citecolor=Violet, linkcolor=red, urlcolor=magenta]{hyperref}       
\usepackage{url}            
\usepackage{booktabs}   

\usepackage{amsfonts}       
\usepackage{nicefrac}       
\usepackage{microtype}      
\usepackage{lipsum}
\usepackage{fancyhdr}       
\usepackage{wrapfig}
\usepackage{colortbl}
\usepackage[dvipsnames]{xcolor}
\usepackage{graphicx}    
\graphicspath{{media/}}     
\usepackage[most]{tcolorbox}
\usepackage{alltt}
\usepackage{enumitem}
\usepackage{bm}
\usepackage{bbm}
\usepackage{tikz}
\usetikzlibrary{arrows.meta}

\usepackage{amsmath}
\usepackage{amssymb}
\usepackage{mathtools}
\usepackage{amsthm}
\usepackage{amsfonts} 
\usepackage{thmtools, thm-restate} 

\usepackage[capitalize,noabbrev]{cleveref}

\usepackage[vlined,ruled,linesnumbered]{algorithm2e}
\usepackage{adjustbox}
\usepackage{multirow,mathtools}

\usepackage{etoc}

\usepackage{threeparttable}
\usepackage{arydshln} 

\theoremstyle{plain}
\newtheorem{theorem}{Theorem}[section]
\newtheorem{proposition}[theorem]{Proposition}

\theoremstyle{definition}

\theoremstyle{remark}

\newtcbox{\hlprimarytab}{on line, box align=base, colback=BlueGreen!20,colframe=blue,size=fbox,arc=3pt, before upper=\strut, top=-2.5pt, bottom=-4.5pt, left=-2pt, right=-2pt, boxrule=0pt}
\newtcbox{\hlsecondarytab}{on line, box align=base, colback=WildStrawberry!10,colframe=orange,size=fbox,arc=3pt, before upper=\strut, top=-2.5pt, bottom=-4.5pt, left=-2pt, right=-2pt, boxrule=0pt}

\newcommand{\refs}[2]{\hyperref[#1]{\ref*{#1}#2}}

\newtcolorbox[auto counter, number within=section]{mydefinition}[2][]{%
  enhanced,
  breakable,
  colback=green!1,
  colframe=green!40!black,
  colbacktitle=green!10!white,
  coltitle=black,
  fonttitle=\bfseries\small,
  fontupper=\small,
  title={Definition~\thetcbcounter: #2},
  label={def:#2},
  attach boxed title to top left={yshift=-3.25mm, xshift=2mm},
  boxed title style={colframe=green!40!black},
  left=10pt,
  right=10pt,
  top=9pt,
  bottom=3pt,
  arc=1.5pt,
  drop shadow=black!25,
  #1
}

\newtcolorbox[auto counter, number within=section]{mythm}[2][]{%
  enhanced,
  breakable,
  colback=cyan!1, 
  colframe=cyan!40!black, 
  colbacktitle=cyan!10!white, 
  coltitle=black, 
  fonttitle=\bfseries\small,
  fontupper=\small, 
  title={Theorem~\thetcbcounter: #2}, 
  label={thm:#2},
  attach boxed title to top left={yshift=-3.25mm, xshift=2mm},
  boxed title style={colframe=cyan!35!black}, 
  left=10pt, 
  right=10pt, 
  top=9pt, 
  bottom=3pt, 
  arc=1.5pt, 
  drop shadow=black!25, 
  #1
}

\usepackage{wrapfig}

\BeforeBeginEnvironment{wrapfigure}{\setlength{\intextsep}{0pt}}
\BeforeBeginEnvironment{wraptable}{\setlength{\intextsep}{0pt}}

\usepackage{comment} 
\usepackage[most]{tcolorbox}
\usepackage{soul}
\usepackage{xcolor}

\newcommand{\ace}{\textcolor{Violet}{\textbf{ACE}}}
\newcommand{\am}{\textcolor{Violet!65}{\textbf{TCE-AM}}}

\newtcolorbox{diagnosticbox}[1][]{
  enhanced, breakable,
  colback=gradblue!2,
  colframe=gradblue!45,
  boxrule=0.45pt,
  leftrule=2.4pt,
  arc=2pt, outer arc=2pt,
  left=6pt, right=6pt, top=3pt, bottom=3pt,
  before skip=3pt, after skip=3pt,
  colback=white,
  colframe=black!16,
  boxrule=0.4pt,
  borderline west={2.6pt}{0pt}{gradblue},
  arc=2pt, outer arc=2pt,
  left=8pt, right=8pt, top=5pt, bottom=6pt,
  before skip=6pt, after skip=6pt,
  fontupper=\small,
  fonttitle=\bfseries\sffamily\small,
  coltitle=gradblue,
  colbacktitle=gradblue!9,
  coltitle=black,
  colbacktitle=white,
  title={#1},
  toptitle=2pt, bottomtitle=2pt,
  lefttitle=5pt, righttitle=5pt,
  segmentation style={solid, gradblue!20, line width=0.35pt},
  toptitle=5pt, bottomtitle=4pt,
  lefttitle=8pt, righttitle=8pt,
  titlerule=0.4pt,
  segmentation style={solid, black!12, line width=0.4pt},
}

\definecolor{qxAce}{RGB}{94,53,150}      
\definecolor{qxTce}{RGB}{178,62,48}      
\definecolor{qxInk}{RGB}{52,56,70}       
\definecolor{qxMute}{RGB}{116,120,134}

\newtcbox{\qxtag}{on line, enhanced, frame hidden, boxrule=0pt, arc=3pt,
  colback=white!86!qxInk, coltext=qxInk, boxsep=0pt,
  left=3.5pt, right=3.5pt, top=1.4pt, bottom=1.4pt,
  fontupper=\sffamily\scriptsize\bfseries}

\newtcolorbox{qxcard}[3][]{enhanced, unbreakable,
  colback=white, colframe=qxInk!35, boxrule=0.5pt, arc=3pt,
  left=7pt, right=7pt, top=6pt, bottom=6pt,
  before skip=8pt, after skip=10pt,
  fontupper=\small,
  title={#2\hfill #3},
  fonttitle=\sffamily\small\bfseries, coltitle=white,
  colbacktitle=qxInk, toptitle=3pt, bottomtitle=3pt, lefttitle=7pt, righttitle=5pt,
  drop fuzzy shadow=black!18, #1}

\newlength{\qxlab}
\newcommand{\qxfield}[2]{\par\noindent
  \makebox[\qxlab][l]{\raisebox{0.4pt}{\sffamily\scriptsize\bfseries\color{qxMute}\MakeUppercase{#1}}}%
  \parbox[t]{\dimexpr\linewidth-\qxlab\relax}{\raggedright\leavevmode #2\strut}\par\vspace{1.5pt}}

\newcommand{\qxrule}{\par\vspace{2pt}\noindent{\color{qxInk!18}\rule{\linewidth}{0.4pt}}\par\vspace{3pt}}

\tcbset{
  qxresp/.style={enhanced, boxrule=0.5pt, arc=2pt,
    left=5pt, right=5pt, top=3.5pt, bottom=4pt, fontupper=\small,
    fonttitle=\sffamily\scriptsize\bfseries, coltitle=white,
    toptitle=1.6pt, bottomtitle=1.6pt, lefttitle=5pt, righttitle=5pt},
  qxsame/.style={colback=black!3, colframe=qxMute!70, colbacktitle=qxMute,
    title={TCE\hfill $\checkmark$\; #1}}}
\newtcolorbox{qxace}[2][]{qxresp, colback=qxAce!5, colframe=qxAce!60,
  colbacktitle=qxAce, title={ACE\hfill $\checkmark$\; #2}, #1}
\newtcolorbox{qxtce}[2][]{qxresp, colback=qxTce!5, colframe=qxTce!60,
  colbacktitle=qxTce, title={TCE\hfill $\boldsymbol{\times}$\; #2}, #1}

\newenvironment{qxpair}{\par\vspace{2pt}%
  \begin{tcolorbox}[blanker, before skip=2pt, after skip=2pt]%
  \begin{tcbraster}[raster columns=5, raster equal height, raster column skip=5pt,
    raster before skip=0pt, raster after skip=0pt]}{\end{tcbraster}\end{tcolorbox}}
\newenvironment{qxstack}{\par\vspace{2pt}%
  \begin{tcolorbox}[blanker, before skip=2pt, after skip=2pt]%
  \begin{tcbraster}[raster columns=1, raster row skip=4pt,
    raster before skip=0pt, raster after skip=0pt]}{\end{tcbraster}\end{tcolorbox}}

\newcommand{\qxhit}[1]{\textcolor{qxAce}{\textbf{#1}}}
\newcommand{\qxmiss}[1]{\textcolor{qxTce}{\textbf{#1}}}
\newcommand{\qxnote}[1]{{\sffamily\scriptsize\color{qxMute}[#1]}}
\newcommand{\qxref}[1]{\unskip\quad{\sffamily\scriptsize\bfseries\color{qxMute}REFERENCE}\;\,\mbox{#1}}

\tcolorboxenvironment{proposition}{
  enhanced,
  breakable,
  colback=white,
  colframe=black!30,
  boxrule=0.6pt,
  arc=2pt,
  outer arc=2pt,
  left=8pt,
  right=8pt,
  top=2pt,
  bottom=2pt,
  before skip=6pt,
  after skip=6pt,
}

\newtheorem*{remark*}{Remark}

\definecolor{gradblue}{RGB}{31,119,180}

\title{Encoding Context into LLM Weights: \\ What Makes It Context?}

\author{Yue Wang$^{1}$\quad Qizhou Wang$^{2}$ \quad Feng Liu$^{1}$  \quad Gang Niu$^{2}$ \quad Masashi Sugiyama$^{2,3}$ \quad Bo Han$^{4}$ \\ \\ $^1$ The University of Melbourne \quad $^2$ RIKEN Center for Advanced Intelligence Project \\  $^3$ The University of Tokyo \quad $^4$ TMLR Group, Hong Kong Baptist University  }

\iclrfinalcopy 
\begin{document}

\etocdepthtag.toc{mtchapter}
\maketitle

\begin{abstract}
Longer contexts allow large language models (LLMs) to condition their predictions on more information, but inference costs grow prohibitively with context length. An emerging alternative uses test-time updates to encode context as temporary changes to model weights. However, encoding this information into weights that already represent general knowledge obscures its role as context, so the LLM may fail to use it properly to condition subsequent inference. To restore this contextual role, we first associate each context with an identifier during encoding, thereby distinguishing the newly encoded information from pre-existing knowledge. Invoking the identifier at inference then enables the LLM to recite the encoded context. Nevertheless, successful recitation does not ensure that the LLM will use the recalled information as context. We therefore further introduce a prompting framework that explicitly guides the LLM to use this information as context. Across diverse controlled tasks and practical benchmarks, our method outperformed the baseline using only off-the-shelf LLMs, showing its advances in context encoding.

\end{abstract}

\section{Introduction}
\label{sec:intro}

{Large language models} (LLMs) have advanced substantially in recent years~\citep{brown2020language,dubey2024llama3,deepseekai2025r1}, and a key foundation of this progress is their expanding capacity to process long contexts~\citep{team2024gemini}  and to scale inference-time computation~\citep{muennighoff2025s1}. The former allows LLMs to ingest larger volumes of diverse inputs, including elaborate prompts, documents, multimodal data, dialogue histories, and tool-interaction records~\citep{openai2023gpt4}. The latter supports test-time scaling through extended reasoning trajectories involving iterative refinement and verification~\citep{wei2022chain,snell2024scaling}.

These benefits, however, come with limitations. As context grows, key-value (KV) cache memory scales linearly and causal attention computation scales quadratically, imposing prohibitive hardware demands and deployment costs~\citep{pope2023efficiently}. {Models may also lose track of relevant information in longer contexts}~\citep{liu2024lost,du2025context}, which has been linked to boundary-token biases~\citep{hsieh2024found}, limited length generalization~\citep{zhong2025understanding}, and attention dispersion~\citep{velivckovic2024softmax}. LLMs are therefore constrained by their {input context windows}, reaching only the million-token scale even in the most advanced models~\citep{yang2501qwen2,openai2026gpt55}, while their effectively usable length can be shorter~\citep{hsieh2024ruler,an2025does}.

\begin{figure}
    \centering
    \includegraphics[width=.84\linewidth]{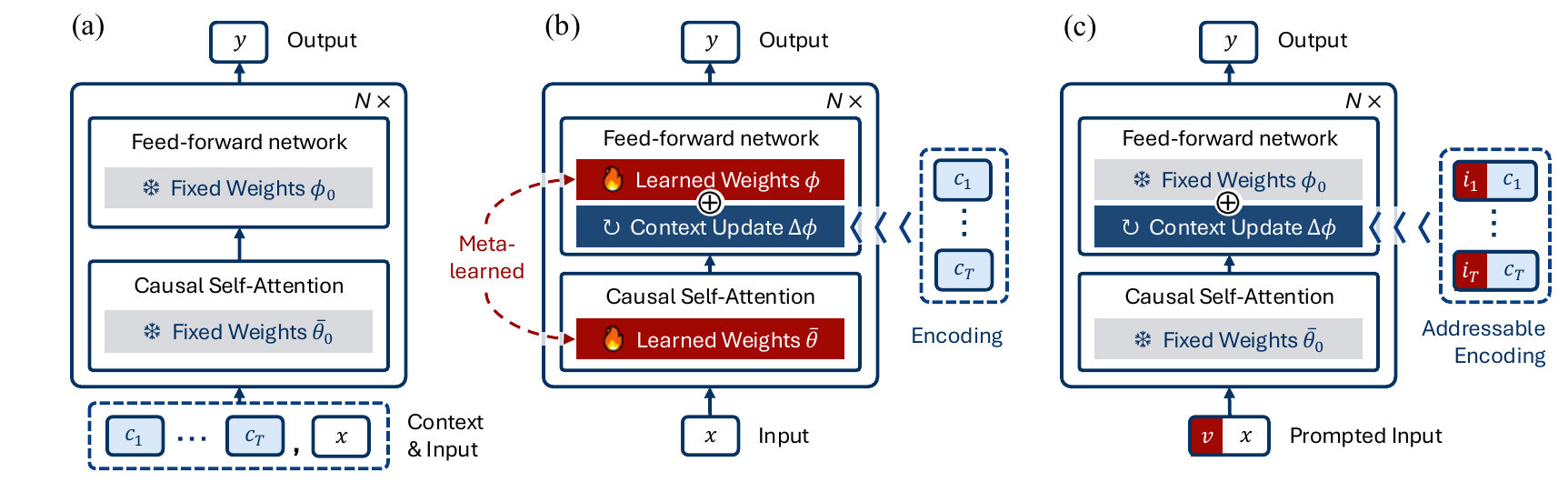}
    \caption{Comparison of (a) full-context inference, (b) encoded-context inference with meta-trained TCE, and (c) encoded-context inference with \ace. Full-context inference jointly processes $c_{1:T}$ and $x$, becoming infeasible when their combined length exceeds the input context window. TCE and \ace\ instead encode the chunks $c_{1:T}$ into LLM weights, illustrated here as FFN weights. Prior TCE methods often rely on additional meta-training-based fine-tuning, whereas \ace\ operates directly on off-the-shelf LLMs without such fine-tuning.}
    \label{fig:Illustrative}
\end{figure}

Taken together, the above observations reveal a tension: LLMs benefit from consuming more tokens at inference, yet an ever-growing context eventually exhausts the available context window. This tension motivates recent work on \emph{test-time context encoding} (TCE)~\citep{chen2023reckoning,sun2024learning,tandon2025end,zhang2025test,behrouz2026titans,feng2026inplace,charakorn2026doc}, which encodes prior context through temporary weight updates rather than retaining it in the input (Figure~\ref{fig:Illustrative}). By encoding context without occupying KV cache positions, TCE can potentially mitigate the computational costs and performance degradation associated with long input contexts.
Ideally, an LLM should condition its predictions on encoded information as reliably as if that information remained within its effective input window. Prior work has reported promising TCE gains using autoregressive encoding~\citep{chen2023reckoning,tandon2025end,chen2025perk}, 
often paired with additional meta-training-based fine-tuning, given its ``\emph{learning to (learn at test time)}'' nature~\citep{sun2024learning}, to improve the use of encoded information.

However, additional meta-training can itself improve downstream performance, making it difficult to attribute the reported gains specifically to context encoding. This leaves open whether TCE can work effectively with off-the-shelf LLMs. 
Although autoregressive objectives can enable memorization~\citep{chen2023reckoning}, our experiments in Section~\ref{sec:exp} showed that TCE without such fine-tuning remained far inferior to \emph{full-context inference}, which retains the context in the input.
This gap highlights the distinction between memorization and contextual use, as also discussed in prior work~\citep{eyuboglu2025cartridges,chen2025perk}. We therefore investigate how off-the-shelf LLMs can reliably use encoded information as context without such task-specific meta-training.

This problem is nevertheless challenging because newly encoded context and pre-existing general knowledge share the same weights, while current TCE objectives provide no explicit signal that distinguishes the encoded information as active context. This lack of distinction, in turn, makes the influence of such information on subsequent inference difficult to control, leaving it uncertain whether the resulting LLM behavior will be reliably conditioned on that context.
To this end, we propose \emph{Addressable Context Encoding} (\ace), which makes encoded context distinguishable from pre-existing knowledge through a simple addressing mechanism. Specifically, during encoding, we associate each context with a \emph{unique identifier} (UID), such as a randomly generated alphanumeric string. At inference, invoking the relevant UIDs directs the LLM to recite the associated content, thereby providing an explicit access path to the encoded information. Moreover, going beyond mere recitation, we wrap this addressing mechanism in a flexible prompting framework~\citep{kojima2022large} that further guides the LLM to selectively use and reason over the encoded information when responding to subsequent requests (Figure~\ref{fig:Illustrative}), thus making its behavior more controllable.

We stress-tested \ace\ on controlled tasks covering various aspects of context use, observing substantial improvements over classical TCE. For example, on two such tasks, Recall and Recognition, \ace\ increased accuracy from $1.5\%$ to $75.6\%$ and from $0.0\%$ to $98.0\%$, respectively, on Qwen3.5-27B~\citep{team2026qwen3}.
These benefits extended to long-context benchmarks, including \textsc{RULER}~\citep{hsieh2024ruler}, widely used in prior TCE studies~\citep{tandon2025end,feng2026inplace}, and the more challenging \textsc{LongBench}~\citep{bai2024longbench} and \textsc{LongBench v2}~\citep{bai2025longbench}, which have received less attention in this setting.
With Muse-Glimmer-30B~\citep{meta2026museglimmer}, \ace\ improved RULER NIAH-3 accuracy from $8\%$ to $63\%$ and LongBench answer F1 from $14.0$ to $19.2$.

\section{Background}\label{sec:bkgd}
Let $p_{\theta_0}$ denote the conditional distribution of a well-trained LLM with weights $\theta_0$ and an $L_{\max}$-token input context window. Let $C$ denote prior context, $x$ a subsequent request, and $y$ the corresponding response. When $C$ and $x$ fit within the window, full-context inference generates
\begin{equation}
    y\sim p_{\theta_0}(\cdot\mid C, x),
    \qquad |C|+|x|\leq L_{\max},\label{eq:plain infer}
\end{equation}
where $|\cdot|$ denotes token length. In this setting, the model can directly use the information and interaction state in $C$ when generating $y$ in response to $x$.

\subsection{Long-context Inference} 
As \(|C|+|x|\) approaches \(L_{\max}\), {full-context inference} incurs increasing memory and compute costs~\citep{dao2022flashattention}, and the model may use relevant information less reliably~\citep{liu2024lost}. Mainstream approaches alleviate these problems through context compression~\citep{jiang2024longllmlingua} or KV-cache eviction~\citep{zhang2023h2o} but risk discarding task-relevant information. \emph{Test-time context encoding} (TCE)~\citep{chen2023reckoning,sun2024learning,tandon2025end,zhang2025test,behrouz2026titans} offers a promising alternative by storing context in a high-capacity \emph{parametric state}, potentially preserving more information than explicit context-reduction methods.

We consider a general long-context setting in which accumulated interaction histories may yield $|C|\gg L_{\max}$, as in long-running agentic deployments~\citep{wu2024longmemeval}. We therefore partition $C$ into $T$ temporally ordered, contiguous chunks $c_{1:T}$ such that
\begin{equation}
    C=c_1\Vert\cdots\Vert c_T,
    \qquad |c_t|\leq B,
\end{equation}
where $\Vert$ denotes concatenation. 
The {chunk budget} $B$ ensures that each encoding input fits within $L_{\max}$. The context $C$ may include factual information, dialogue history, instructions, preferences, pending queries, or unfinished reasoning, while $x$ may be a query, instruction, or continuation cue.

\subsection{Test-time Context Encoding} 
\label{sec:bkgd, tce}
A subset of the model weights, referred to as \emph{fast weights}~\citep{feng2026inplace}, serves as the parametric state into which the context is encoded. 
We write $\theta_0=(\bar{\theta}_0,\phi_0)$, where $\phi_0$ denotes the initial fast weights, to which we reset them before encoding each independent context $C$, while $\bar{\theta}_0$ remains frozen during TCE. 
For each chunk $c_t$, we minimize the autoregressive loss with respect to $\phi$, i.e.,
\begin{equation}
    \mathcal{L}(\phi;c_t)
    = -\log p_{\bar{\theta}_0,\phi}(c_t).
    \label{eq:encode}
\end{equation}
Starting from $\phi_0$, we encode the chunks $c_1,\ldots,c_T$ in temporal order. Let $\phi_t^{(k)}$ denote the fast weights after $k$ optimization steps on the chunk $c_t$. We initialize $\phi_t^{(0)}=\phi_{t-1}$ from the state produced by the preceding chunks and perform $K$ updates as follows:
\begin{equation}
    \phi_t^{(k+1)} = \operatorname{Update}\Big(\phi_t^{(k)},\nabla_{\phi}\mathcal{L}(\phi_t^{(k)};c_t)\Big),\qquad k=0,\ldots,K-1,\qquad \phi_t=\phi_t^{(K)}.\label{eq:encode-update}
\end{equation}
$\operatorname{Update}$ denotes one optimizer step, such as gradient descent~\citep{robbins1951stochastic} or Adam~\citep{kingma2014adam}. 
Thus, $\phi_t$ accumulates updates from the ordered context prefix $c_{1:t}$, yielding the context-encoded model weights
$\theta_C=(\bar{\theta}_0,\phi_T)$ with respect to $C$ after all $T$ chunks.

Once encoded, $C$ is removed from the input, and the context-encoded LLM generates
\begin{equation}
    y\sim p_{\theta_C}(\cdot\mid x),
    \qquad |x|\leq L_{\max}.\label{eq:infer}
\end{equation}
Information from $C$ is expected to influence subsequent predictions through the updated fast weights $\phi_T$. The goal is for the resulting LLM to use this information as reliably as it would if $C$ remained directly available in the input~\citep{chen2023reckoning,eyuboglu2025cartridges,chen2025perk}.
When $C$ is no longer needed, such as after a conversation ends, its context-specific updates can be discarded by resetting the fast weights from $\phi_T$ to $\phi_0$, thereby restoring the LLM weights from $\theta_C$ to $\theta_0$.

\subsection{Scope of This Study}
\label{sec:scope}
Prior work has explored TCE objectives other than the autoregressive loss~\citep{sun2024learning,behrouz2026titans,feng2026inplace,charakorn2026doc}, as discussed in Section~\ref{sec:related works}. However, these alternatives may perform worse than autoregressive encoding~\citep{tandon2025end}, generalize poorly to practical use~\citep{caccia2025training}, or incur substantial computational costs~\citep{behrouz2026titans}. We therefore primarily focus on autoregressive TCE.

Most prior work reporting promising results relies on additional meta-training~\citep{chen2023reckoning,behrouz2026titans,tandon2025end,feng2026inplace,kuratov2026gradmem,chen2025perk}. 
Such fine-tuning may introduce reliance on task-specific correlations, making it difficult to separate its contribution from that of the underlying TCE mechanism, as we discuss later.
We therefore restrict our study to off-the-shelf LLMs, aiming to establish a clearer understanding of TCE reliability.

Moreover, TCE requires backpropagation and thus incurs additional computation. Nevertheless, fast-weight updates are typically restricted to selected LLM weights~\citep{tandon2025end,chen2025perk,feng2026inplace}, reducing the number of trainable weights. 
Further improvements in efficiency and implementation are beyond our focus on reliability and are left for future work.

\section{New Method}\label{sec:newmethods}

Without task-specific meta-training, the TCE framework introduced in Section~\ref{sec:bkgd, tce} {performed far worse than full-context inference, even in relatively simple experiments with very short contexts} (Section~\ref{sec:controlled-sandbox}). We attribute these failures in part to the fact that autoregressive encoding updates LLM weights much like pretraining and supervised fine-tuning, without explicitly identifying the encoded information as context. 
Although prior work reports a tendency for weight updates to bias subsequent predictions toward newly encoded information~\citep{meng2022locating,malladi2023kernel,zhang2025uncovering}, this implicit bias does not reliably ensure that the LLM recalls the encoded information, let alone treats it as context.
These observations suggest two challenges: \textbf{(C1)} {making encoded information reliably accessible} and \textbf{(C2)} specifying how it should function as context.

Similar views have been raised in prior work~\citep{chen2023reckoning,tandon2025end,chen2025perk}, yet these approaches do not resolve the fundamental limitation but instead turn to task-specific meta-training to sidestep it.
We fill this gap by tackling \textbf{C1} and \textbf{C2} through two components, respectively: an addressable mechanism for making encoded information reliably accessible and explicitly distinguishing it from pre-existing LLM knowledge (Section~\ref{sec:ace-addressable}), and a prompting framework for specifying how it should function as context, thus guiding its proper use in subsequent predictions (Section~\ref{sec:prompting}). 
Combining these two components, we propose \emph{Addressable Context Encoding} (\ace), an enhanced TCE method that uses only off-the-shelf LLMs, as detailed in Section~\ref{sec:overall method}.

\subsection{Addressable Mechanism}
\label{sec:ace-addressable}

Enabling LLMs to reliably use encoded context first requires making the encoded information explicitly distinguishable from pre-existing knowledge, as specified by \textbf{C1}. One way to pursue this goal is to make the encoded context \emph{addressable} by binding the context to a \emph{unique identifier} (UID) during encoding. Conditioning on the UID provides an explicit cue for accessing the corresponding information during inference, thereby helping distinguish it from other knowledge.

Formally, we assign each context chunk $c_t$ a UID $i_t$. For example, in \texttt{ctx-573277c6}, \texttt{ctx-} indicates context and the suffix is a random alphanumeric string. This format is illustrative, and we impose no particular format or design rule beyond requiring uniqueness. 
To associate each chunk with its UID, we modify the autoregressive objective in Eq.~\eqref{eq:encode} to predict the chunk conditioned on its UID, i.e.,
\begin{equation}
    \mathcal{L}_{\mathrm{A}}(\phi;c_t,i_t) = -\log p_{\bar{\theta}_0,\phi}(c_t\mid i_t),
    \label{eq:address encode}
\end{equation}
which learns an association between the UID $i_t$ and its context chunk $c_t$. We retain the sequential fast-weight updates as in Eq.~\eqref{eq:encode-update}, replacing only the basic autoregressive loss with $\mathcal{L}_{\mathrm{A}}$. 

At inference, we remove the original context and supply the corresponding UIDs in temporal order. Let $I_C=i_1\Vert\cdots\Vert i_T$ denote this ordered sequence. Given a subsequent input $x$, the model generates
\begin{equation}
    y\sim p_{\theta_C}(\cdot\mid I_C,x),
    \qquad |I_C|+|x|\leq L_{\max}.
    \label{eq:ace-addressable-inference}
\end{equation}
Since each UID is short, the added token overhead is small for a moderate number of chunks, leaving most of the input budget for the new request $x$. 
$I_C$ references the encoded context without reproducing its content in the input, providing an explicit cue for the LLM to recall the encoded information and potentially facilitating its use in responding to the request $x$.
Appendix~\ref{app:addressable-analysis} further provides mathematical insights into selective access to encoded context and a proof-of-concept analysis showing how UID conditioning can improve encoding dynamics.

\subsection{Prompting Enhancement}\label{sec:prompting}

Through this addressing mechanism, the LLM can recall the corresponding context when prompted with its UID. When further given a request $x$, it is expected to prioritize relevant encoded information over pre-existing knowledge. 
However, this behavior depends on inherent model capabilities that may vary across models and are difficult to characterize, leaving implicit how the encoded information should guide subsequent inference. This motivates explicitly specifying its contextual role and use, as required by \textbf{C2}.
To address this limitation, we build a prompting framework on top of the addressing mechanism. It leverages the instruction-following capabilities of off-the-shelf LLMs to explicitly specify the role and use of encoded information as context during both encoding and inference.

{During encoding}, we make the role and intended use of each context chunk $c_t$ explicit in natural language through a \emph{role-binding prompt}. It identifies the chunk as prior conversational context, specifies its temporal order, highlights its assigned UID, and indicates that relevant information should be used when that UID is invoked. We use the following template:

\begin{diagnosticbox}[Role-Binding Prompt]
\textit{The following text is a context chunk from the
current conversation at $s_t$, identified by $i_t$.
It has been encoded into the model and should be treated
as conversational context whenever $i_t$ is invoked:
$c_t$}
\end{diagnosticbox}

Here $s_t$ denotes the timestamp assigned to chunk $c_t$, indicating its temporal order. These instructions clarify the contextual role of the chunk and provide metadata about its identity and temporal order. To encode this information alongside the chunk, we introduce an additional autoregressive objective that complements the UID-conditioned objective in Eq.~\eqref{eq:address encode}. Specifically, let $b_t$ denote the prompt prefix preceding $c_t$ in the template above. We define the role-binding objective as
\begin{equation}
\mathcal{L}_{\mathrm B}(\phi;c_t,b_t)
=-\log p_{\bar{\theta}_0,\phi}(b_t\Vert c_t).
\label{eq:role-bind}
\end{equation}
During inference, we offer a more direct prompt to guide the use of encoded information.  As in the addressing mechanism, we supply the UIDs $I_C$ of the encoded chunks $C$, but further instruct the model to selectively use their associated content. Specifically, we ask the LLM to recall only information relevant to the current request $x$ and use it to formulate the response, without reciting the entire context, which would otherwise consume the limited context window and undermine the original purpose of TCE. We introduce the \emph{context-invocation prompt}, which is given by
\begin{diagnosticbox}[Context-Invocation Prompt (\texttt{ci\_v1})]
\textit{Parts of the conversation have been encoded as the context chunks, ordered by time: $i_1$ at $s_1$, $\ldots$, $i_T$ at $s_T$. Treat these chunks as prior conversational context in subsequent inference. Selectively use information relevant to the current request without reproducing unrelated details.}
\end{diagnosticbox}
We label this template \texttt{ci\_v1} to distinguish it from other templates evaluated in our experiments and shown in Appendix~\ref{app:examples-prompts}. We denote the \emph{context-invocation prompt} for the encoded context $C$ with UIDs $I_C$ by $v_C$. Then, given a subsequent input $x$, we prepend $v_C$ and generate
\begin{equation}
    y\sim p_{\theta_C}(\cdot\mid v_C,x),
    \label{eq:contt-infer}
\end{equation}
providing more explicit control over the use of encoded information than the addressing mechanism alone. Note that we still assume that the context window is large enough to accommodate $v_C$ and $x$.

\subsection{Overall Algorithm}
\label{sec:overall method}
We summarize the overall \ace\ algorithm, which consists of encoding and inference procedures. During encoding, the model sequentially encodes the context chunks. Starting from $\phi_t^{(0)}=\phi_{t-1}$, we perform the following update for $k=0,\ldots,K-1$:
\begin{equation}
\phi_t^{(k+1)}
= \operatorname{Update}\left(
\phi_t^{(k)},
\left.
\nabla_{\phi}\left[
\mathcal{L}_{\mathrm A}(\phi;c_t,i_t)
+\mathcal{L}_{\mathrm B}(\phi;c_t,b_t)
\right]
\right|_{\phi=\phi_t^{(k)}}
\right).
\end{equation}
After $K$ steps, we set $\phi_t=\phi_t^{(K)}$. Once all chunks have been encoded, we obtain $\theta_C=(\bar{\theta}_0,\phi_T)$. Then, given $x$, the encoded LLM generates $y\sim p_{\theta_C}(\cdot\mid v_C,x)$. Unlike full-context inference, $v_C$ does not reproduce the original chunks. Instead, it provides a compact index of the encoded chunks and specifies how they should guide inference. Moreover, when the context $C$ is no longer needed, we restore the original model weights $\theta_0$ by resetting the fast weights from $\phi_T$ to $\phi_0$.

\section{Experiments}\label{sec:exp}
We first evaluated \ace\ in controlled experiments covering different aspects of context use. We then assessed its practical effectiveness on standard long-context benchmarks: \textsc{RULER}~\citep{hsieh2024ruler}, widely used in prior work~\citep{tandon2025end,feng2026inplace}, and \textsc{LongBench}~\citep{bai2024longbench} and \textsc{LongBench v2}~\citep{bai2025longbench}, which have received less attention in prior work.
\subsection{Sandbox Tasks}
\label{sec:controlled-sandbox}

To begin with, we constructed a controlled benchmark to
stress-test TCE across different forms of context use and
examine whether \ace\ improved its reliability.
The benchmark used template-based synthesis and factual
data curation, pairing short contexts with queries.
The contexts averaged approximately $11$ tokens under the
Qwen3.5 tokenizer. We targeted $200$ instances for each
context-use task or subtask and removed invalid candidates
through further screening. We also included a separately
constructed Retention test. These tasks are briefly described below.

\begin{itemize}[leftmargin=1.2em, itemsep=2pt, topsep=0pt]
\item \textbf{Recall.} We constructed $197$ synthetic entity--attribute--value statements, such as ``\emph{The registry code of Korvustre Wharf is GC-483.}'' Each served as context for a recall query that provided the entity and attribute to request the value, e.g., ``\emph{What is the registry code of Korvustre Wharf?}'' A response was scored as correct if it retrieved the right value.

\item \textbf{Recognition.}
Using the same $197$ statements as in Recall, we asked, ``\emph{What information were you just told?}'', which relies on recency as the only cue in the request. A response was scored as correct if it retrieved not only the correct value, but also the correct entity and attribute. 
Recall and Recognition distinguish content-cued retrieval from identifying recently encoded information.

\item \textbf{Continuation.}
We created $168$ self-contained factual and arithmetic questions with short, automatically verifiable answers. Each question served as the context, and the LLM was asked to continue the conversation by answering the unresolved question. 
We measured accuracy, evaluating whether the model can recognize an encoded question as a pending request and act on it.

\item \textbf{Conflict.}
We tested conflicts between newly encoded context and pre-existing knowledge across three subtasks: Counterfactual ($200$ instances), Dispute ($43$ instances), and Outdated Information ($17$, $20$, and $33$ instances for Qwen3.5-4B, 9B, and 27B, respectively). These data were model-specific because pre-existing knowledge varies across models. 
By default, we measured accuracy against the contextual answer rather than that from pre-existing knowledge.

\item \textbf{Retention.} 
As an additional check, we evaluated accuracy on $100$ random query instances unrelated to the encoded context, covering arithmetic, general knowledge, language, and temporal reasoning. This aimed to assess whether TCE impaired general LLM performance.
\end{itemize}

{We offer illustrative examples of our curated sandbox datasets in Appendix}~\ref{app:examples-datasets}. The context of each instance forms a single encoding chunk, i.e., $T=1$, because it is sufficiently short. Note that all these tasks are relatively easy for full-context inference, which achieves near-perfect performance, but remain challenging for TCE, as we will demonstrate later.

By default, we updated only the weights of the final feed-forward network (FFN) layer, following prior work on selective FFN updates~\citep{mitchell2021fast,tandon2025end,feng2026inplace} and motivated by its interpretation as a knowledge store~\citep{geva2021transformer,dai2022knowledge,meng2022locating}. We used Adam~\citep{kingma2014adam} with an initial learning rate of $10^{-4}$ and gradient clipping with a maximum norm of $1$~\citep{pascanu2013difficulty}, without further hyperparameter tuning. Moreover, we adopted a stopping threshold $\tau$, set to $0.1$ by default, and stopped encoding when the maximum context-token loss fell below $\tau$ for $\mathcal{L}_{\rm B}$ in \ace\ or after $256$ update steps, whichever occurred first. Beyond the context-invocation prompt introduced in Section~\ref{sec:prompting}, termed \texttt{ci\_v1}, we evaluated two other variants, \texttt{ci\_v2} and \texttt{ci\_short}, to examine the impact of prompting on the reliability of \ace. {Details of all three prompts can be found in Appendix}~\ref{app:examples-prompts}.

We first evaluated and compared classical TCE discussed in Section~\ref{sec:bkgd, tce} and our proposed enhancement \ace\ on the Recall, Recognition, and Continuation tasks. We also reported results on Retention as a supplement to assess the impact of context encoding on general model performance. We tested both methods on the 4B, 9B, and 27B versions of the Qwen3.5 model family.
Table~\ref{tab:sandbox-retrieval-use} reports accuracy results. As shown, classical TCE fails to reliably use encoded context even on the relatively easy Recall task. In contrast, \ace\ substantially improves performance across the context-use tasks, making context encoding more usable with off-the-shelf LLMs. Larger LLMs do not necessarily perform better, which is plausible given that these models were not specifically trained for TCE. 

Performance also depends on the invocation prompt: \texttt{ci\_short} achieves the best results in many cases, possibly because the contexts are short and the tasks are relatively simple.
This advantage does not consistently extend to more challenging benchmarks such as \textsc{RULER}, as shown in Appendix~\ref{app:more long context}, highlighting the importance of appropriate prompting~\citep{du2025context}.
Finally, with only the final FFN serving as fast weights, both TCE and \ace\ maintain high accuracy on the Retention task, suggesting that the evaluated general capabilities remain largely intact after encoding.

\begin{table*}[t]
\caption{Comparison between TCE and \ace\ on Recall, Recognition, Continuation, and Retention. We reported accuracy (\%) on the Qwen3.5 family. Bold marks the best result within each model and task.}
\label{tab:sandbox-retrieval-use}
\centering
\small
\setlength{\tabcolsep}{5.5pt}
\begin{tabular}{ccccccc}
\toprule
LLM & Method & Prompt & Recall & Recognition & Continuation & Retention \\
\midrule
\multirow{4}{*}{Qwen3.5-4B}
 & TCE  & --                         & 6.1 & 0.5 & 0.0 & \textbf{100.0} \\
 & \ace & \texttt{ci\_short} & 52.3 & \textbf{91.9} & 46.4 & 99.0 \\
 & \ace & \texttt{ci\_v1}    & 55.3 & 81.7 & \textbf{48.8} & \textbf{100.0} \\
 & \ace & \texttt{ci\_v2}    & \textbf{60.4} & 80.7 & 28.0 & \textbf{100.0} \\
\midrule
\multirow{4}{*}{Qwen3.5-9B}
 & TCE  & --                         & 4.6 & 2.0 & 1.2 & \textbf{100.0} \\
 & \ace & \texttt{ci\_short} & \textbf{40.1} & \textbf{96.4} & 9.5 & \textbf{100.0} \\
 & \ace & \texttt{ci\_v1}    & 35.5 & 95.4 & \textbf{13.7} & \textbf{100.0} \\
 & \ace & \texttt{ci\_v2}    & 38.1 & 94.4 & 5.4 & \textbf{100.0} \\
\midrule
\multirow{4}{*}{Qwen3.5-27B}
 & TCE  & --                         & 1.5 & 0.0 & 0.0 & \textbf{100.0} \\
 & \ace & \texttt{ci\_short} & \textbf{75.6} & \textbf{98.0} & \textbf{20.8} & \textbf{100.0} \\
 & \ace & \texttt{ci\_v1}    & 39.1 & 58.4 & 11.9 & \textbf{100.0} \\
 & \ace & \texttt{ci\_v2}    & 45.2 & 50.8 & 2.4 & \textbf{100.0} \\
\bottomrule
\end{tabular}
\end{table*}

\begin{table*}[t]
\caption{Comparison between TCE and \ace\ on the three categories of
Conflict: Counterfactuals (Counterfact), Disputes (Dispute),
and Outdated Information (Outdated). We further reported two-sided
Counterfactual accuracy in the column labeled
``Counterfact (two-sided).'' }
\label{tab:sandbox-conflict}
\centering
\small
\setlength{\tabcolsep}{5.5pt}
\begin{tabular}{ccccccc}
\toprule
\multirow{2}{*}{LLM}
& \multirow{2}{*}{Method}
& \multirow{2}{*}{Prompt}
& \multirow{2}{*}{Counterfact}
& \multirow{2}{*}{Dispute}
& \multirow{2}{*}{Outdated}
& Counterfact \\
& & & & & & (two-sided) \\
\midrule
\multirow{4}{*}{Qwen3.5-4B}
& TCE  & --                & 3.5 & 7.0 & 5.9 & 0.5 \\
& \ace & \texttt{ci\_short} & 86.5 & 83.7 & 88.2 & 20.0 \\
& \ace & \texttt{ci\_v1}    & 92.5 & \textbf{95.3} & \textbf{100.0} & 27.5 \\
& \ace & \texttt{ci\_v2}    & \textbf{93.0} & 90.7 & 94.1 & \textbf{28.0} \\
\midrule
\multirow{4}{*}{Qwen3.5-9B}
& TCE  & --                & 14.0 & 11.6 & 10.0 & 6.0 \\
& \ace & \texttt{ci\_short} & 94.0 & 95.3 & \textbf{100.0} & 29.0 \\
& \ace & \texttt{ci\_v1}    & 94.5 & 95.3 & \textbf{100.0} & \textbf{30.5} \\
& \ace & \texttt{ci\_v2}    & \textbf{95.0} & \textbf{97.7} & \textbf{100.0} & \textbf{30.5} \\
\midrule
\multirow{4}{*}{Qwen3.5-27B}
& TCE  & --                & 11.0 & 18.6 & 54.5 & 8.0 \\
& \ace & \texttt{ci\_short} & \textbf{96.0} & \textbf{97.7} & 97.0 & \textbf{36.5} \\
& \ace & \texttt{ci\_v1}    & 93.0 & 88.4 & \textbf{100.0} & 29.0 \\
& \ace & \texttt{ci\_v2}    & 92.0 & 93.0 & 97.0 & 31.5 \\
\bottomrule
\end{tabular}
\end{table*}

Table~\ref{tab:sandbox-conflict} reports results separately for the Counterfactual, Dispute, and Outdated Information categories of Conflict. We further reported two-sided accuracy for the Counterfactual category, where an instance was scored as correct only if the LLM correctly answered both queries: one using the encoded context and the other explicitly asking it to disregard that context and use pre-existing knowledge instead. 
Interestingly, TCE sometimes performed better on Conflict than on the tasks in Table~\ref{tab:sandbox-retrieval-use}, and \ace\ even achieved near-perfect accuracy in several settings. This suggests potential for resolving knowledge conflicts, though further validation is needed. However, both methods performed worse on two-sided Counterfactual, which requires query-dependent source selection, but \ace\ still substantially outperformed TCE.
Taken together, these results suggest that TCE struggles to distinguish encoded context from pre-existing knowledge, while \ace\ offers a promising way to mitigate this issue. However, there remains room for improvement, which may motivate further research.

\begin{table*}[t]
\caption{Comparison between TCE and \ace\ on \textsc{RULER}, \textsc{LongBench}, and \textsc{LongBench v2}. We reported retrieval accuracy (\%) on \textsc{RULER}, answer F1 ($\times100$) on \textsc{LongBench}, and summed judge-assigned scores over $100$ instances on \textsc{LongBench v2}. By default, we use \texttt{ci\_v2}. Gold Standard refers to full-context inference with the context retained in the input. Bold marks the better result between TCE and \ace.}
\label{tab:long-context}
\centering
\small
\setlength{\tabcolsep}{5pt}
\begin{tabular}{ccccccc}
\toprule
\multirow{2}{*}{LLM}
& \multirow{2}{*}{Method}
& \multicolumn{3}{c}{\textsc{RULER}}
& \multirow{2}{*}{\textsc{LongBench}}
& \multirow{2}{*}{\textsc{LongBench v2}} \\
\cmidrule(lr){3-5}
& & NIAH-1 & NIAH-2 & NIAH-3 & & \\
\midrule
\multirow{3}{*}{Qwen3.5-4B}
 & Gold Standard & 100 & 100 & 100 & 20.8 & 2.5 \\
\cmidrule(lr){2-7}
 & TCE  & 98 & 7 & 0 & 8.8 & 0.0 \\
 & \ace & \textbf{100} & \textbf{35} & \textbf{15}
        & \textbf{9.0} & \textbf{2.0} \\
\midrule
\multirow{3}{*}{Qwen3.5-9B}
 & Gold Standard & 100 & 100 & 100 & 20.3 & 3.0 \\
\cmidrule(lr){2-7}
 & TCE  & 88 & 7 & 1 & 8.2 & 2.5 \\
 & \ace & \textbf{100} & \textbf{17} & \textbf{27}
        & \textbf{9.6} & \textbf{3.5} \\
\midrule
\multirow{3}{*}{Qwen3.5-27B}
 & Gold Standard & 100 & 100 & 100 & 18.9 & 4.0 \\
\cmidrule(lr){2-7}
 & TCE  & 85 & 7 & 0 & 7.9 & \textbf{2.5} \\
 & \ace & \textbf{100} & \textbf{40} & \textbf{29}
        & \textbf{9.8} & \textbf{2.5} \\
\midrule
\multirow{3}{*}{Muse-Glimmer-30B}
 & Gold Standard & 100 & 100 & 100 & 21.2 & 3.0 \\
\cmidrule(lr){2-7}
 & TCE  & 27 & 33 & 8 & 14.0 & 2.0 \\
 & \ace & \textbf{89} & \textbf{43} & \textbf{63}
        & \textbf{19.2} & \textbf{3.0} \\
\bottomrule
\end{tabular}
\end{table*}

\subsection{Long-context Tasks}
\label{sec:long-context-evaluation}

Finally, we evaluated the effectiveness of \ace\ on long-context benchmarks to assess whether its benefits extend beyond sandbox settings. In addition to the Qwen3.5 family, we included Muse-Glimmer-30B~\citep{meta2026museglimmer} to examine the generalizability across model families. Alongside \textsc{RULER}~\citep{hsieh2024ruler}, which was also used in prior work, we considered two more realistic and challenging long-context benchmarks, \textsc{LongBench} and \textsc{LongBench v2}~\citep{bai2024longbench,bai2025longbench}. We briefly introduce them before presenting the results.
\begin{itemize}[leftmargin=1.2em, itemsep=2pt, topsep=0pt]
\item \textbf{RULER.} 
We evaluated three \emph{Needle-in-a-Haystack} (NIAH) tasks, NIAH-1, NIAH-2, and NIAH-3, from \textsc{RULER}, using $100$ instances per task.
They require the model to retrieve a target value embedded in irrelevant text, with increasing difficulty through more realistic distractors and more complex target values. 
In our experiments, each instance used a $1{,}024$-token context, and we measured performance by the accuracy of retrieving the needle.

\item \textbf{LongBench.}
We evaluated $100$ document question-answering instances on \textsc{LongBench}, using its length-balanced version. These questions require the model to locate and use relevant information in the supplied documents. The contexts ranged from $2$k to $7$k tokens, and we measured answer F1, which captures token overlap between the generated and ground-truth answers.

\item \textbf{LongBench v2.} 
We considered \textsc{LongBench v2} as a more challenging benchmark requiring deeper understanding and reasoning over longer and more complex contexts. We evaluated $100$ instances with contexts ranging from about $10$k to $15$k tokens.
These tasks require integrating information across documents, understanding detailed document content, or learning from examples provided in context, posing challenges even for full-context inference. 
GPT-5.5~\citep{openai2026gpt55} judged responses given the question, candidate answers, and gold answer, with method identities withheld. It assigned $1$, $0.5$, or $0$ points to correct, borderline, or incorrect responses, respectively. 
\end{itemize}

Most experimental settings and hyperparameters were the same as in the sandbox experiments, including Adam, the learning rate, and gradient clipping.
However, to accommodate longer contexts, we adopted a maximum chunk size of $1{,}024$ tokens and encoded the chunks sequentially. For example, a $7k$-token context was divided into seven chunks. For \textsc{LongBench} and \textsc{LongBench v2}, we set $\tau=8$ rather than $0.1$ to limit encoding time and reduce the risk of overfitting.

Table~\ref{tab:long-context} reports all results. On the relatively simple NIAH-1 task in \textsc{RULER}, even TCE performed well with the Qwen3.5 models. However, its accuracy dropped sharply on the harder NIAH variants, while \ace\ always improved over TCE across models and tasks. Performance nevertheless remained far from perfect on NIAH-2 and NIAH-3, demonstrating both the benefits of \ace\ and room for further improvement.
On \textsc{LongBench} and \textsc{LongBench v2}, performance remained limited even with full-context inference, reflecting the difficulty of these tasks. Nevertheless, \ace\ improved over TCE on \textsc{LongBench} and matched or exceeded it on \textsc{LongBench v2}. 

\section{Related Work}
\label{sec:related works}
Our work falls broadly within \emph{test-time training} (TTT)~\citep{sun2020test}, which updates model weights at test time to adapt to distribution shifts, as studied in the broader domain-adaptation literature~\citep{sugiyama2007covariate,ben2010theory}. Recent work has extended this paradigm to LLMs~\citep{hardt2024test,bansal2026let,zhu2026self}, typically by optimizing an autoregressive loss over the available context.
However, these methods primarily adapt the model to condition more effectively on the available context while retaining the original context or its KV cache during subsequent generation, which differs from our setup.

Going beyond this setting, more recent TTT methods use weight updates to replace the input context entirely~\citep{chen2023reckoning}, closely matching our context-encoding setup for long-context inference. This setting is substantially more challenging than domain adaptation because the model must faithfully retain the information in its weights and use this implicit representation to condition subsequent inference. This lies beyond conventional TTT methods~\citep{lu2026locas,song2026beyond}, and enhanced methods reporting strong performance after context removal typically rely on task-specific meta-training~\citep{chen2023reckoning,chen2025perk}. 
This makes it unclear whether the gains reflect general context encoding or adaptation to the meta-training distribution, and, to our knowledge, no ablation study has formally examined this distinction.

Beyond the autoregressive loss, prior work has explored alternative TTT objectives under the same setting, with the input context removed after encoding. One prominent direction uses KV association objectives~\citep{sun2024learning,zhang2025test,behrouz2025atlas,behrouz2026titans,tang2026modular,feng2026inplace}, which encode context as a bias for subsequent queries, guiding outputs toward values associated with relevant contextual keys. 
Despite their simple update rules, KV association inherently optimizes token matching rather than information encoding, potentially increasing reliance on meta-training and typically underperforming autoregressive encoding~\citep{tandon2025end,liu2026test}. 
Another direction distills full-context behavior into a model operating without the context~\citep{caccia2025training,zheng2026context}. 
Although less reliant on further fine-tuning, it requires representative context-query pairs spanning diverse, unpredictable user requests and optimization over substantially more test-time data, raising practical cost and efficiency concerns.

\section{Summary}
TCE offers a promising approach to long-context inference by encoding context into model weights instead of retaining it in the input. Prior work often relies on additional task-specific meta-training, making it difficult to distinguish general context-encoding capability from adaptation to the meta-training distribution. We therefore studied TCE with only off-the-shelf LLMs. Our results highlight that encoding information into model weights does not ensure its reliable use as context, limiting the practical effectiveness of TCE.
To address this limitation, we proposed \ace, which combines UID-conditioned encoding for explicit access to encoded information with a prompting framework that further specifies its contextual role and intended use. Experiments on controlled sandbox tasks and long-context benchmarks demonstrated improvements over classical TCE. These findings establish \ace\ as a promising step toward reliable context encoding, while highlighting substantial room for future improvement, particularly in source selection and more complex long-context reasoning.







\bibliography{iclr2026_conference}
\bibliographystyle{iclr2026_conference}
\clearpage

\appendix

\section{Dataset Examples and Prompt Templates}

We provide examples and prompt templates to clarify the
experimental setup. The dataset examples illustrate the
encoded contexts, subsequent queries, and expected answers
for each sandbox task and subtask. We then present the three
context-invocation prompts used to guide the models in
accessing and using encoded information during inference.

\subsection{Sandbox Dataset Examples}
\label{app:examples-datasets}

To give a more concrete picture of the sandbox datasets, we provide an example for each task and subtask: Recall, Recognition, Continuation, the three Conflict categories (Counterfactual, Dispute, and Outdated Information), and finally Retention. The Outdated Information example comes from the Qwen3.5-4B dataset, with values based on the Wikidata snapshot collected on August 22, 2026. We retain the original wording ``context above'' in the Conflict queries, although the context itself is removed from the input after encoding.

\begin{diagnosticbox}[Recall and Recognition]
\textbf{Context:}
The registry code of Korvustre Wharf is GC-483.

\medskip
\textbf{Recall query:}
What is the registry code of Korvustre Wharf?
Return only the value.

\textbf{Reference answer:}
GC-483

\medskip
\textbf{Recognition query:}
What information were you just told?
Return the complete statement.

\textbf{Reference fields:}
Entity: Korvustre Wharf;
attribute: registry code;
value: GC-483.
\end{diagnosticbox}

\begin{diagnosticbox}[Continuation]
\textbf{Factual context:}
What is the capital of Norway?

\textbf{Reference answer:}
Oslo

\medskip
\textbf{Arithmetic context:}
What is 26 squared?

\textbf{Reference answer:}
676

\medskip
\textbf{Subsequent request for either context:}
Continue the preceding conversation. It ended with an
unanswered question, which is stored in the encoded
context that is available to you. Recall that question
and give your answer to it now. Do not reply that you
lack access to it. End your reply with a line of the
form ``Final answer: \textless answer\textgreater''.
\end{diagnosticbox}

\begin{diagnosticbox}[Counterfactual]
\textbf{Context:}
The capital of Peru is Arequipa.

\medskip
\textbf{Contextual-answer query:}
Answer using only the context above, even if it
contradicts what you believe to be true.
According to that context, what is the capital of Peru?
Return only the value.

\textbf{Reference answer:}
Arequipa

\medskip
\textbf{Prior-knowledge query:}
Setting aside the current context, what is the
generally known capital of Peru?
Return only the value.

\textbf{Reference answer:}
Lima
\end{diagnosticbox}

\begin{diagnosticbox}[Dispute]
\textbf{Context:}
The inventor of the telephone is Antonio Meucci.

\medskip
\textbf{Contextual-answer query:}
Answer using only the context above, even if it
contradicts what you believe to be true.
According to that context, what is the inventor
of the telephone? Return only the value.

\textbf{Reference answer:}
Antonio Meucci
\end{diagnosticbox}

\begin{diagnosticbox}[Outdated Information]
\textbf{Context:}
The head coach of FC Bayern Munich is Vincent Kompany.

\medskip
\textbf{Contextual-answer query:}
Answer using only the context above, even if it
contradicts what you believe to be true.
According to that context, what is the head coach
of FC Bayern Munich? Return only the value.

\textbf{Reference answer:}
Vincent Kompany

\medskip
\textbf{Outdated value:}
Thomas Tuchel
\end{diagnosticbox}

\begin{diagnosticbox}[Retention]
\textbf{Encoded context:}
The registry code of Trilbane Outpost is RV-807.

\medskip
\textbf{Unrelated question:}
What is 17 multiplied by 4?
Return only the number.

\textbf{Reference answer:}
68

\medskip
\textbf{Unrelated question:}
What is the capital of Portugal?
Return only the value.

\textbf{Reference answer:}
Lisbon
\end{diagnosticbox}

\subsection{Context-Invocation Prompts}
\label{app:examples-prompts}

In addition to the context-invocation prompt introduced in Section~\ref{sec:prompting}, denoted by \texttt{ci\_v1}, we evaluated two variants while keeping the encoding procedure and role-binding prompt fixed. The short variant, \texttt{ci\_short}, uses a minimal instruction to invoke the encoded context. The extended variant, \texttt{ci\_v2}, additionally instructs the model to continue an unresolved request or incomplete task in the latest chunk.
We provide all three prompts below. Let $i_t$ and $s_t$ denote the UID and timestamp of context chunk $c_t$, respectively. Each prompt is followed by a blank line and the task query, and the resulting text is supplied as a user message.

\begin{diagnosticbox}[Context-Invocation Prompt (\texttt{ci\_short})]
\textit{Recall context $i_1,\ldots,i_T$ to continue.}
\end{diagnosticbox}

\begin{diagnosticbox}[Context-Invocation Prompt (\texttt{ci\_v1})]
\textit{Parts of the conversation have been encoded
as the context chunks, ordered by time:
$i_1$ at $s_1$, $\ldots$, $i_T$ at $s_T$.
Treat these chunks as prior conversational context
in subsequent inference. Selectively use information
relevant to the current request without reproducing
unrelated details.}
\end{diagnosticbox}

\begin{diagnosticbox}[Context-Invocation Prompt (\texttt{ci\_v2})]
\textit{Parts of the conversation have been encoded
as the context chunks, ordered by time:
$i_1$ at $s_1$, $\ldots$, $i_T$ at $s_T$.
Treat these chunks as prior conversational context
in subsequent inference. Selectively use information
relevant to the current request without reproducing
unrelated details. If the latest chunk $i_T$ contains
an unresolved request or incomplete task, continue
from it when appropriate.}
\end{diagnosticbox}

\clearpage
\section{Mathematical Insights}
\label{app:addressable-analysis}

We analyze \ace\ from two complementary perspectives. {For inference}, Appendix~\ref{app:addressing-benefit1} shows how UIDs enable the model to distinguish among encoded contexts with addressable encoding. Appendix~\ref{app:source-distinction} extends this analysis by showing how UIDs provide an explicit reference for selecting encoded context, thereby supporting its distinction from pre-existing knowledge.
For encoding, Appendix~\ref{app:addressing-learning} provides a proof-of-concept analysis showing that UID conditioning can strengthen gradients and accelerate learning. Together, these results illustrate the potential benefits of UID conditioning for both generation and encoding.

\subsection{Context Distinction}
\label{app:addressing-benefit1}

We discuss whether conditioning on UIDs increases the probability
of generating encoded context relative to generation without UIDs,
thereby supporting an explicit mechanism for context distinction.
We compare the TCE objectives $\mathcal{L}$ and
$\mathcal{L}_{\mathrm A}$, introduced in
Sections~\ref{sec:bkgd, tce} and~\ref{sec:ace-addressable},
respectively, first at their ideal optima and then under
finite encoding error.

We consider an LLM that receives contexts $c_1,\ldots,c_T$
from $T$ distinct sources, respectively, with each context
$c_t$ assigned a unique UID $i_t$. We reuse the context-chunk
notation, but consider independent contexts rather than
chunks of a single context. This simplifies the mathematical
modeling of context use: all contexts are encoded into the
same model weights, and one context is selected at inference.
For this analysis, we assume that the contexts are pairwise
distinct and each contains $B$ tokens. Generation distributions
are defined over length-$B$ sequences.

We evaluate generation by sampling a context according to
a distribution $\pi=(\pi_1,\ldots,\pi_T)$, where $\pi_t>0$
is the probability of selecting $c_t$ and
$\sum_{t=1}^{T}\pi_t=1$. We assume the same context distribution
during encoding and evaluation, with no distribution mismatch.
Let $\theta_C$ and $\theta_C^{\mathrm A}$ denote the weights
after TCE without and with UID conditioning, obtained using
the encoding objectives
\[
\mathcal{L}(\theta)
:=-\sum_{t=1}^{T}\pi_t\log p_\theta(c_t),
\qquad
\mathcal{L}_{\mathrm A}(\theta)
:=-\sum_{t=1}^{T}\pi_t\log p_\theta(c_t\mid i_t),
\]
respectively.
We suppress dependence on $\pi$, $c_{1:T}$, and $i_{1:T}$,
and write $p(c):=p_{\theta_C}(c)$ and
$p_{\mathrm A}(c\mid i_t):=p_{\theta_C^{\mathrm A}}(c\mid i_t)$
for brevity.
The same objectives measure generation performance after
encoding and can be expressed as
\[
\mathcal{L}(p)
:=-\sum_{t=1}^{T}\pi_t\log p(c_t),
\qquad
\mathcal{L}_{\mathrm A}(p_{\mathrm A})
:=-\sum_{t=1}^{T}\pi_t\log p_{\mathrm A}(c_t\mid i_t).
\]
These objectives measure the expected negative log-likelihood
of generating a target context sampled according to $\pi$.
Lower values indicate better fit to the target contexts,
and $\mathcal{L}(p)-\mathcal{L}_{\mathrm A}(p_{\mathrm A})$
quantifies the generation benefit of UID-based context selection.

We first consider the ideal case in which both objectives
are minimized over unrestricted distributions.

\begin{proposition}
\label{prop:optimal-addressing}
Over unrestricted distributions, the optimal generation
probabilities satisfy
\[
p^*(c_t)=\pi_t,
\qquad
p_{\mathrm A}^*(c_t\mid i_t)=1
\]
for every $t$. Consequently,
\[
\frac{p_{\mathrm A}^*(c_t\mid i_t)}{p^*(c_t)}
=\frac{1}{\pi_t},
\qquad
\inf_p\mathcal{L}(p)
-\inf_{p_{\mathrm A}}\mathcal{L}_{\mathrm A}(p_{\mathrm A})
=H(\pi),
\]
where $H(\pi):=-\sum_{t=1}^{T}\pi_t\log\pi_t$ is the entropy
of the context-sampling distribution.
For uniform weights, the probability ratio is $T$ for every
context, and the optimal loss gap is $\log T$.
\end{proposition}

\begin{proof}
Define the weighted empirical distribution over contexts by
\[
\bar q(c):=\sum_{t=1}^{T}\pi_t\mathbf{1}\{c=c_t\}.
\]
Since the contexts are pairwise distinct,
$\bar q(c_t)=\pi_t$ for every $t$.
The cross-entropy decomposition gives
\[
D_{\mathrm{KL}}(\bar q\Vert p)
=\sum_{t=1}^{T}\pi_t\log\frac{\pi_t}{p(c_t)}=\mathcal{L}(p)-H(\pi).
\]
By the nonnegativity of KL divergence,
\[
\mathcal{L}(p)
=H(\pi)+D_{\mathrm{KL}}(\bar q\Vert p)
\geq H(\pi).
\]
The lower bound is attained at $p^*=\bar q$, yielding
$p^*(c_t)=\pi_t$ and $\inf_p\mathcal{L}(p)=H(\pi)$.

The UID-conditioned objective is nonnegative.
Because the UIDs are unique, each conditional distribution
can assign probability one to its corresponding context:
\[
p_{\mathrm A}^*(c\mid i_t)=\mathbf{1}\{c=c_t\}.
\]
This gives $\mathcal{L}_{\mathrm A}(p_{\mathrm A}^*)=0$.
Conversely, since every $\pi_t>0$, zero loss requires
$p_{\mathrm A}^*(c_t\mid i_t)=1$ for every $t$.
Combining the two optima gives the stated probability ratio
and loss gap. For uniform weights, $\pi_t=1/T$ and
$H(\pi)=\log T$.
\end{proof}

\begin{remark*}
UIDs provide an explicit input signal that identifies which
encoded context is being requested.
Without UID conditioning, the model must allocate probability
across the encoded contexts, whereas conditioning on each UID
allows it to assign probability one to the corresponding
context.
\end{remark*}

Proposition~\ref{prop:optimal-addressing} discusses the case
in which both encoding objectives reach their optima.
In practice, encoding may leave residual error.
We therefore examine whether UID conditioning still provides
an average generation advantage when neither model is assumed
optimal. The following result shows that this advantage is
guaranteed when the UID-conditioned encoding error is
sufficiently small.

\begin{proposition}
\label{prop:finite-addressing}
Suppose $p(c_t)>0$ and $p_{\mathrm A}(c_t\mid i_t)>0$
for every $t=1,\ldots,T$.
Let $\epsilon_{\mathrm A}:=\mathcal{L}_{\mathrm A}(p_{\mathrm A})$
denote the final UID-conditioned encoding error. Then
\[
\mathcal{L}(p)-\mathcal{L}_{\mathrm A}(p_{\mathrm A})
\geq H(\pi)-\epsilon_{\mathrm A}.
\]
Equivalently,
\[
\prod_{t=1}^{T}
\left[
\frac{p_{\mathrm A}(c_t\mid i_t)}{p(c_t)}
\right]^{\pi_t}
\geq \exp\bigl(H(\pi)-\epsilon_{\mathrm A}\bigr).
\]
Thus, $\epsilon_{\mathrm A}<H(\pi)$ guarantees a positive
average log-probability gain.
For uniform weights, the lower bounds are
$\log T-\epsilon_{\mathrm A}$ and
$T\exp(-\epsilon_{\mathrm A})$, respectively, and the
sufficient condition becomes $\epsilon_{\mathrm A}<\log T$.
\end{proposition}

\begin{proof}
Using the cross-entropy decomposition from the proof of
Proposition~\ref{prop:optimal-addressing}, we obtain
\[
\mathcal{L}(p)-\mathcal{L}_{\mathrm A}(p_{\mathrm A})
=H(\pi)+D_{\mathrm{KL}}(\bar q\Vert p)
-\epsilon_{\mathrm A}\geq H(\pi)-\epsilon_{\mathrm A},
\]
where the inequality follows from the nonnegativity of
KL divergence.

By the definitions of the objectives,
\[
\mathcal{L}(p)-\mathcal{L}_{\mathrm A}(p_{\mathrm A})
=\sum_{t=1}^{T}\pi_t
\log\frac{p_{\mathrm A}(c_t\mid i_t)}{p(c_t)}=\log\prod_{t=1}^{T}
\left[
\frac{p_{\mathrm A}(c_t\mid i_t)}{p(c_t)}
\right]^{\pi_t}.
\]
Exponentiating the loss-gap bound yields the probability-ratio
bound. If $\epsilon_{\mathrm A}<H(\pi)$, the loss gap is
strictly positive. Substituting $H(\pi)=\log T$ gives the
uniform-weight results.
\end{proof}

\begin{remark*}
The sufficient condition does not require perfect encoding:
a nonzero final encoding error below $H(\pi)$ suffices to
guarantee an average generation advantage.
This guarantee concerns selection among multiple encoded
contexts; for $T=1$, the ideal loss gap is zero.
\end{remark*}

\subsection{Source Distinction}
\label{app:source-distinction}

We extend Proposition~\ref{prop:finite-addressing} to
include pre-existing knowledge as an additional source,
examining whether UID conditioning helps distinguish
encoded context from pre-existing knowledge.
Let $\widetilde C$ denote a length-$B$ target sequence
with its realization $\tilde c$, and let
$Z\in\{0,1,\ldots,T\}$ denote its source.
Here, $Z=0$ represents pre-existing knowledge, while $Z=t$
represents encoded context $c_t$.
With $\alpha\in(0,1)$ denoting the probability of selecting
pre-existing knowledge as the target source, we define
\[
\Pr(Z=0)=\alpha,
\qquad
\Pr(Z=t)=(1-\alpha)\pi_t,
\quad t=1,\ldots,T.
\]
Let $q_K$ denote a fixed empirical distribution over
length-$B$ sequences expressing pre-existing knowledge.
The target distributions are
\[
q(\tilde c\mid Z=0)=q_K(\tilde c),
\qquad
q(\tilde c\mid Z=t)=\mathbf{1}\{\tilde c=c_t\}.
\]
Together with the source probabilities, these define a joint
distribution $q$ over $(\widetilde C,Z)$, with marginal
\[
q(\tilde c)
=\alpha q_K(\tilde c)+(1-\alpha)\bar q(\tilde c),
\]
where $\bar q$ is the weighted empirical distribution over
encoded contexts defined in
Appendix~\ref{app:addressing-benefit1}.

We retain the notation $p$ and $p_{\mathrm A}$ from
Appendix~\ref{app:addressing-benefit1}.
For encoded contexts, the UID-conditioned model receives
the corresponding UID $i_t$.
For targets from pre-existing knowledge, it receives no UID.
For brevity, we write
$p_{\mathrm A}(\tilde c):=p_{\theta_C^{\mathrm A}}(\tilde c)$.
Thus, $p_{\mathrm A}(\tilde c)$ and
$p_{\mathrm A}(\tilde c\mid i_t)$ refer to the same
UID-conditioned model without and with a UID, respectively,
whereas $p(\tilde c)$ refers to the baseline encoded model.

The corresponding weighted negative log-likelihood objectives
are
\[
\begin{aligned}
\mathcal J(p)
&:=-\alpha\sum_{\tilde c:q_K(\tilde c)>0}
q_K(\tilde c)\log p(\tilde c)
+(1-\alpha)\mathcal L(p),\\
\mathcal J_{\mathrm A}(p_{\mathrm A})
&:=-\alpha\sum_{\tilde c:q_K(\tilde c)>0}
q_K(\tilde c)\log p_{\mathrm A}(\tilde c)
+(1-\alpha)\mathcal L_{\mathrm A}(p_{\mathrm A}).
\end{aligned}
\]
These objectives evaluate generation from pre-existing knowledge
without a UID and from encoded context with or without its UID,
according to the encoding method.

\begin{proposition}
\label{prop:source-addressing}
Suppose $p(\tilde c)>0$ whenever $q(\tilde c)>0$,
$p_{\mathrm A}(\tilde c)>0$ whenever $q_K(\tilde c)>0$,
and $p_{\mathrm A}(c_t\mid i_t)>0$ for every $t$.
Define the excess loss as
\[
\epsilon_{\mathrm A}^{\mathrm{src}}
:=\mathcal J_{\mathrm A}(p_{\mathrm A})
-H(\widetilde C\mid Z).
\]
Then
\[
\mathcal J(p)-\mathcal J_{\mathrm A}(p_{\mathrm A})
\geq I(\widetilde C;Z)-\epsilon_{\mathrm A}^{\mathrm{src}},
\]
where $I(\widetilde C;Z)$ is the mutual information between
the target sequence and its source under $q$.
Equivalently,
\[
\prod_{\tilde c:q_K(\tilde c)>0}
\left[
\frac{p_{\mathrm A}(\tilde c)}{p(\tilde c)}
\right]^{\alpha q_K(\tilde c)}
\prod_{t=1}^{T}
\left[
\frac{p_{\mathrm A}(c_t\mid i_t)}{p(c_t)}
\right]^{(1-\alpha)\pi_t}\geq
\exp\bigl(I(\widetilde C;Z)-\epsilon_{\mathrm A}^{\mathrm{src}}\bigr).
\]
Thus, $\epsilon_{\mathrm A}^{\mathrm{src}}<I(\widetilde C;Z)$
is sufficient for a positive average log-probability gain
under this evaluation.
\end{proposition}

\begin{proof}
Since
$\mathcal J(p)=-\sum_{\tilde c:q(\tilde c)>0}
q(\tilde c)\log p(\tilde c)$,
the cross-entropy decomposition gives
\[
\mathcal J(p)
=H(\widetilde C)+D_{\mathrm{KL}}(q\Vert p),
\]
where $q$ in the KL divergence denotes the marginal
distribution of $\widetilde C$.
Since the target context is deterministic given $Z=t$
for every $t\geq1$, we have
\[
H(\widetilde C\mid Z)=\alpha H(q_K),
\]
where
$H(q_K):=-\sum_{\tilde c:q_K(\tilde c)>0}
q_K(\tilde c)\log q_K(\tilde c)$.
Consequently,
\[
\epsilon_{\mathrm A}^{\mathrm{src}}
=\mathcal J_{\mathrm A}(p_{\mathrm A})-\alpha H(q_K)=\alpha D_{\mathrm{KL}}(q_K\Vert p_{\mathrm A})
+(1-\alpha)\mathcal L_{\mathrm A}(p_{\mathrm A})
\geq0,
\]
where $p_{\mathrm A}$ in the KL divergence denotes the
model distribution without a UID.

Substituting the definition of
$\epsilon_{\mathrm A}^{\mathrm{src}}$ yields
\[
\begin{aligned}
\mathcal J(p)-\mathcal J_{\mathrm A}(p_{\mathrm A})
=H(\widetilde C)-H(\widetilde C\mid Z)
+D_{\mathrm{KL}}(q\Vert p)
-\epsilon_{\mathrm A}^{\mathrm{src}}
&=I(\widetilde C;Z)+D_{\mathrm{KL}}(q\Vert p)
-\epsilon_{\mathrm A}^{\mathrm{src}}\\
&\geq I(\widetilde C;Z)-\epsilon_{\mathrm A}^{\mathrm{src}}.
\end{aligned}
\]
By the definitions of the objectives,
\[
\mathcal J(p)-\mathcal J_{\mathrm A}(p_{\mathrm A})
=\alpha\sum_{\tilde c:q_K(\tilde c)>0}
q_K(\tilde c)\log
\frac{p_{\mathrm A}(\tilde c)}{p(\tilde c)}\quad+(1-\alpha)\sum_{t=1}^{T}\pi_t
\log\frac{p_{\mathrm A}(c_t\mid i_t)}{p(c_t)}.
\]
Exponentiating the loss-gap bound therefore yields the
probability-ratio bound.
If $\epsilon_{\mathrm A}^{\mathrm{src}}<I(\widetilde C;Z)$,
the loss gap is strictly positive, establishing the average
log-probability advantage.
\end{proof}

\begin{remark*}
The bound characterizes the average generation benefit of
providing UIDs for encoded contexts while using no UID for
pre-existing knowledge.
The excess loss combines two errors:
\[
\epsilon_{\mathrm A}^{\mathrm{src}}
=\alpha D_{\mathrm{KL}}(q_K\Vert p_{\mathrm A})
+(1-\alpha)\epsilon_{\mathrm A}.
\]
If the model remains well fitted to pre-existing knowledge
after encoding, the first term is small.
Provided this term is below $I(\widetilde C;Z)$,
sufficiently low UID-conditioned encoding error guarantees
an average generation advantage under the specified evaluation.
The information term includes both distinctions among encoded
contexts and their distinction from pre-existing knowledge.
However, the bound does not guarantee reliable source selection
for arbitrary downstream requests.
\end{remark*}

\subsection{Encoding Benefits}
\label{app:addressing-learning}

The preceding analysis characterizes the generation benefit
of addressable context. We now examine its encoding benefits.
Prior empirical work associates stronger gradient contributions
with stronger memorization~\citep{huang2022detecting,maini2023can}.
This motivates us to examine whether UID conditioning produces
larger gradients and whether these lead to faster loss reduction.

We focus on encoding the $n$-th token $c_{t,n}$ of context
chunk $c_t$, given its preceding tokens, where $n\geq2$.
We consider a simplified single-head causal attention model
that predicts $c_{t,n}$ from the output at the final prefix
position.
Let $a_1,\ldots,a_{n-1}\in\mathbb R^d$ denote fixed
representations of the preceding tokens.
Without UID conditioning, their attention-weighted
representation is
\[
h:=\sum_{k=1}^{n-1}\omega_k a_k,
\qquad
\omega_k>0,
\qquad
\sum_{k=1}^{n-1}\omega_k=1.
\]
Here, $\omega_k$ is the attention weight assigned to the
$k$-th preceding token when predicting $c_{t,n}$.

We represent the UID $i_t$ by a single token with fixed
representation $v\in\mathbb R^d$.
When the UID is prepended, we assume that it receives
attention weight $\beta\in(0,1)$ while the relative attention
weights among the original tokens remain unchanged.
The aggregated representation then becomes
\[
h_{\mathrm A}:=(1-\beta)h+\beta v.
\]
This corresponds to adding the UID to softmax attention
without changing the attention scores of the original tokens.
For simplicity, we hold the representations and attention
weights fixed throughout training.

The outputs without and with UID conditioning are
\[
z:=W_Vh,
\qquad
z_{\mathrm A}:=W_V^{\mathrm A}h_{\mathrm A},
\]
where $W_V,W_V^{\mathrm A}\in\mathbb R^{M\times d}$ are
the corresponding trainable value projection matrices,
and $M\geq2$ is the vocabulary size.
For simplicity, we take these outputs directly as vocabulary
logits. The token-level negative log-likelihoods are
\[
\ell(W_V)
:=-\log[\operatorname{softmax}(z)]_{c_{t,n}},
\qquad
\ell_{\mathrm A}(W_V^{\mathrm A})
:=-\log[\operatorname{softmax}(z_{\mathrm A})]_{c_{t,n}}.
\]
We update only the value projection matrices and compare
their gradients and loss dynamics.

We identify sufficient conditions under which UID
conditioning increases the norm of the aggregated
representation, and then relate this increase to larger
gradients and faster loss reduction.

\begin{proposition}
\label{prop:addressing-gradient}
Suppose
\[
\langle h,v\rangle=0,
\qquad
0<r:=\frac{\|h\|}{\|v\|}<1.
\]
Then
\[
\gamma:=
\frac{\|h_{\mathrm A}\|^2}{\|h\|^2}
=(1-\beta)^2+\frac{\beta^2}{r^2}.
\]
If, in addition,
\[
\beta>\frac{2r^2}{1+r^2},
\]
then $\gamma>1$.
At identical prediction distributions, i.e.,
$\operatorname{softmax}(z_{\mathrm A})
=\operatorname{softmax}(z)$,
the gradients satisfy
\[
\|\nabla_{W_V^{\mathrm A}}\ell_{\mathrm A}\|_F^2
=\gamma\|\nabla_{W_V}\ell\|_F^2
>\|\nabla_{W_V}\ell\|_F^2.
\]

Under these conditions, consider training both matrices
by gradient flow:
\[
\frac{dW_V}{ds}=-\nabla_{W_V}\ell,
\qquad
\frac{dW_V^{\mathrm A}}{ds}
=-\nabla_{W_V^{\mathrm A}}\ell_{\mathrm A},
\]
where $s\geq0$ denotes training time.
If the initial logits are finite and satisfy
$z_{\mathrm A}(0)=z(0)$, then
\[
z_{\mathrm A}(s)=z(\gamma s).
\]
Consequently, for every $s>0$,
\[
\ell_{\mathrm A}(W_V^{\mathrm A}(s))
=\ell(W_V(\gamma s))
<\ell(W_V(s)).
\]
Thus, UID conditioning accelerates the same loss trajectory
by a factor of $\gamma$.
\end{proposition}

\begin{proof}
Since $h_{\mathrm A}=(1-\beta)h+\beta v$ and
$\langle h,v\rangle=0$,
\[
\|h_{\mathrm A}\|^2
=(1-\beta)^2\|h\|^2+\beta^2\|v\|^2.
\]
Dividing by $\|h\|^2$ gives
\[
\gamma=(1-\beta)^2+\frac{\beta^2}{r^2}.
\]
Under the assumed lower bound on $\beta$,
\[
\gamma-1
=\frac{\beta}{r^2}
\bigl[\beta(1+r^2)-2r^2\bigr]>0.
\]

Let $e_{c_{t,n}}\in\mathbb R^M$ denote the one-hot vector
for the target token.
Differentiating the losses gives
\[
\nabla_{W_V}\ell
=\bigl(\operatorname{softmax}(z)-e_{c_{t,n}}\bigr)h^\top,
\]
\[
\nabla_{W_V^{\mathrm A}}\ell_{\mathrm A}
=\bigl(\operatorname{softmax}(z_{\mathrm A})
-e_{c_{t,n}}\bigr)h_{\mathrm A}^\top.
\]
Using $\|ab^\top\|_F^2=\|a\|^2\|b\|^2$, we obtain
\[
\|\nabla_{W_V}\ell\|_F^2
=\|\operatorname{softmax}(z)-e_{c_{t,n}}\|^2\|h\|^2,
\]
\[
\|\nabla_{W_V^{\mathrm A}}\ell_{\mathrm A}\|_F^2
=\|\operatorname{softmax}(z_{\mathrm A})-e_{c_{t,n}}\|^2
\|h_{\mathrm A}\|^2.
\]
At identical prediction distributions, the prediction-error
factors coincide. Hence,
\[
\|\nabla_{W_V^{\mathrm A}}\ell_{\mathrm A}\|_F^2
=\gamma\|\nabla_{W_V}\ell\|_F^2.
\]
The baseline gradient is nonzero because $h\neq0$ and,
for $M\geq2$, finite logits cannot yield a one-hot softmax
distribution. Together with $\gamma>1$, this establishes
the strict gradient inequality.

Under gradient flow,
\[
\frac{d\ell}{ds}
=-\|\nabla_{W_V}\ell\|_F^2,
\qquad
\frac{d\ell_{\mathrm A}}{ds}
=-\|\nabla_{W_V^{\mathrm A}}\ell_{\mathrm A}\|_F^2.
\]
Thus, at identical predictions, the larger squared gradient
norm yields a loss-decrease rate that is $\gamma$ times
as large.

To compare the full trajectories, recall that $h$ and
$h_{\mathrm A}$ are fixed. The logit dynamics are
\[
\frac{dz}{ds}
=\frac{dW_V}{ds}h
=-\|h\|^2
\bigl(\operatorname{softmax}(z)-e_{c_{t,n}}\bigr),
\]
\[
\frac{dz_{\mathrm A}}{ds}
=-\|h_{\mathrm A}\|^2
\bigl(\operatorname{softmax}(z_{\mathrm A})-e_{c_{t,n}}\bigr).
\]
Since $\|h_{\mathrm A}\|^2=\gamma\|h\|^2$,
the rescaled trajectory $z(\gamma s)$ satisfies the same
differential equation as $z_{\mathrm A}(s)$.
The softmax vector field is globally Lipschitz, so identical
initial logits imply
\[
z_{\mathrm A}(s)=z(\gamma s).
\]
Therefore,
\[
\ell_{\mathrm A}(W_V^{\mathrm A}(s))
=\ell(W_V(\gamma s)).
\]

The logit derivatives are bounded, so finite initial logits
remain finite at every finite time.
The baseline gradient therefore remains nonzero, implying
$d\ell/ds<0$.
Since $\gamma>1$, for every $s>0$,
\[
\ell(W_V(\gamma s))<\ell(W_V(s)),
\]
which completes the proof.
\end{proof}

\begin{remark*}
Under the stated conditions, UID conditioning produces larger
gradients and faster loss reduction, suggesting a potential
memorization benefit in light of prior empirical
findings.
These conditions require the UID representation to be
distinct from the context and exert sufficient influence
on the prediction.
The result is a proof of concept under simplified training
dynamics, rather than a guarantee for full LLM training.
\end{remark*}

\clearpage
\section{Ablation Studies on Sandbox Tasks}
\label{sec:ablations}

Now, we conducted ablations on the sandbox benchmark, focusing on the relatively challenging Recall, Recognition, and Continuation tasks, to examine the contribution of each design choice.

\subsection{Effect of the Addressing Mechanism}

We first compared TCE augmented with the addressing mechanism, termed \am\ for simplicity, against the original TCE. Table~\ref{tab:tce vs am} compares two variants of \am: one follows Section~\ref{sec:ace-addressable}, using Eq.~\eqref{eq:address encode} for encoding and Eq.~\eqref{eq:ace-addressable-inference} for inference (Cue: UID); the other uses the same encoding objective but omits the UID at inference, following Eq.~\eqref{eq:infer} as in TCE (Cue: --).
Overall, \am\ outperformed TCE, with some exceptions on Recall. 
Even without the UID at inference, TCE-AM improved Recognition on 4B and 9B, although its effect on Recall was mixed.
This is consistent with our theoretical analysis in Appendix~\ref{app:addressing-learning}, which suggests that UID-conditioned encoding has the potential to strengthen memorization. 
However, as discussed in Section~\ref{sec:prompting}, the intended use of the encoded information remains implicit, leaving context use unreliable. By further incorporating the prompting framework, \ace\ achieves substantially better performance.

\begin{table*}[t]
\caption{Comparison between TCE, \am, and \ace\ on Recall and
Recognition. We reported accuracy (\%) on the Qwen3.5 family.
Bold marks the best result among TCE and \am\ within each model
and task.}
\label{tab:tce vs am}
\centering
\small
\setlength{\tabcolsep}{5.5pt}
\begin{tabular}{ccccc}
\toprule
LLM & Method & Cue & Recall & Recognition \\
\midrule
\multirow{4}{*}{Qwen3.5-4B}
 & TCE  & --                 & 6.1 & 0.5 \\
 & \am  & --                 & \textbf{7.1} & 10.7 \\
 & \am  & UID                & 5.6 & \textbf{35.5} \\
\cmidrule(lr){2-5}
 & \ace & \texttt{ci\_short} & 52.3 & 91.9 \\
\midrule
\multirow{4}{*}{Qwen3.5-9B}
 & TCE  & --                 & \textbf{4.6} & 2.0 \\
 & \am  & --                 & 1.5 & 3.6 \\
 & \am  & UID                & 0.5 & \textbf{11.2} \\
\cmidrule(lr){2-5}
 & \ace & \texttt{ci\_short} & 40.1 & 96.4 \\
\midrule
\multirow{4}{*}{Qwen3.5-27B}
 & TCE  & --                 & 1.5 & 0.0 \\
 & \am  & --                 & 1.5 & 0.0 \\
 & \am  & UID                & \textbf{4.6} & \textbf{1.0} \\
\cmidrule(lr){2-5}
 & \ace & \texttt{ci\_short} & 75.6 & 98.0 \\
\bottomrule
\end{tabular}
\end{table*}

\subsection{Contributions of Encoding and Invocation}

Table~\ref{tab:component-ablation} reports additional ablations of individual components and their combinations. As shown, removing either $\mathcal{L}_{\mathrm A}$ or $\mathcal{L}_{\mathrm B}$ during encoding reduced performance, highlighting their complementary contributions.
Without encoding, \ace\ scored zero on both tasks, confirming that the invocation prompt alone is insufficient without context-specific weight updates. Using the original TCE inference in Eq.~\eqref{eq:infer} after \ace\ encoding still achieved strong performance in some cases, suggesting that encoded information can remain accessible without an explicit invocation prompt, although the full method performed better. Together, these ablations support the rationale behind our design choices and demonstrate the contribution of each component to reliable context encoding.

\begin{table*}[t]
\caption{Ablation of \ace\ on Recall and Recognition. We reported accuracy (\%) on the Qwen3.5 family. Bold marks the best result within each model and task.}
\label{tab:component-ablation}
\centering
\small
\setlength{\tabcolsep}{5.5pt}
\begin{tabular}{cccc}
\toprule
LLM & Method & Recall & Recognition \\
\midrule
\multirow{5}{*}{Qwen3.5-4B}
 & \ace\ w/o $\mathcal{L}_{\mathrm A}$ & 4.1 & 21.3 \\
 & \ace\ w/o $\mathcal{L}_{\mathrm B}$ & 37.1 & 49.7 \\
 & \ace\ with TCE inference          & 32.5 & 81.2 \\
 & \ace\ w/o encoding                & 0.0 & 0.0 \\
\cmidrule(lr){2-4}
 & \ace & \textbf{55.3} & \textbf{81.7} \\
\midrule
\multirow{5}{*}{Qwen3.5-9B}
 & \ace\ w/o $\mathcal{L}_{\mathrm A}$ & 7.6 & 17.3 \\
 & \ace\ w/o $\mathcal{L}_{\mathrm B}$ & 22.8 & 47.2 \\
 & \ace\ with TCE inference          & 19.8 & 68.0 \\
 & \ace\ w/o encoding                & 0.0 & 0.0 \\
\cmidrule(lr){2-4}
 & \ace & \textbf{35.5} & \textbf{95.4} \\
\midrule
\multirow{5}{*}{Qwen3.5-27B}
 & \ace\ w/o $\mathcal{L}_{\mathrm A}$ & 22.3 & 0.5 \\
 & \ace\ w/o $\mathcal{L}_{\mathrm B}$ & 31.5 & 13.2 \\
 & \ace\ with TCE inference          & 18.3 & 20.8 \\
 & \ace\ w/o encoding                & 0.0 & 0.0 \\
\cmidrule(lr){2-4}
 & \ace & \textbf{39.1} & \textbf{58.4} \\
\bottomrule
\end{tabular}
\end{table*}

\begin{table*}[t]
\caption{Comparison between TCE and \ace\ with different
choices of fast weights on Qwen3.5-4B.
FFN-$k$ and ATTN-$k$ denote the FFN and attention sublayers
of the final $k$ transformer layers, respectively.
We reported accuracy (\%) with \ace\
using \texttt{ci\_v1}.
Bold marks the best result within each method and task.}
\label{tab:fast-weights-4b}
\centering
\small
\setlength{\tabcolsep}{5.5pt}
\begin{tabular}{cccccccc}
\toprule
Method & Fast Weights & Recall & Recognition & Continuation
& Retention & Counterfact & Dispute \\
\midrule
\multirow{6}{*}{TCE}
& FFN-1  & 6.1 & 0.5 & 0.0 & \textbf{100.0} & 3.5 & \textbf{7.0} \\
& FFN-2  & 4.1 & \textbf{7.6} & 0.0 & 68.0 & 3.0 & \textbf{7.0} \\
& FFN-3  & 9.1 & 4.1 & \textbf{3.6} & 43.0 & \textbf{9.5} & 4.7 \\
& ATTN-1 & 5.6 & 0.0 & 0.0 & 84.0 & 0.5 & 2.3 \\
& ATTN-2 & \textbf{11.7} & 0.0 & 0.0 & 74.0 & 1.0 & 2.3 \\
& ATTN-3 & 9.6 & 0.0 & 0.0 & 69.0 & 2.5 & 2.3 \\
\midrule
\multirow{6}{*}{\ace}
& FFN-1  & 55.3 & 81.7 & 48.8 & \textbf{100.0} & 92.5 & 95.3 \\
& FFN-2  & 63.5 & 91.9 & 44.0 & 98.0 & 93.0 & 93.0 \\
& FFN-3  & 55.3 & 89.8 & 52.4 & 95.0 & 87.0 & 90.7 \\
& ATTN-1 & 49.2 & 42.6 & 38.1 & \textbf{100.0} & 93.5 & \textbf{100.0} \\
& ATTN-2 & \textbf{65.0} & 88.8 & 70.2 & \textbf{100.0} & \textbf{96.5} & 97.7 \\
& ATTN-3 & 64.5 & \textbf{92.9} & \textbf{72.0} & \textbf{100.0} & 95.5 & 93.0 \\
\bottomrule
\end{tabular}
\end{table*}

\begin{table*}[t]
\caption{Comparison between TCE and \ace\ with different
choices of fast weights on Qwen3.5-9B.
FFN-$k$ and ATTN-$k$ denote the FFN and attention sublayers
of the final $k$ transformer layers, respectively.
We reported accuracy (\%) with \ace\
using \texttt{ci\_v1}.
Bold marks the best result within each method and task.}
\label{tab:fast-weights-9b}
\centering
\small
\setlength{\tabcolsep}{5.5pt}
\begin{tabular}{cccccccc}
\toprule
Method & Fast Weights & Recall & Recognition & Continuation
& Retention & Counterfact & Dispute \\
\midrule
\multirow{6}{*}{TCE}
& FFN-1  & 4.6 & \textbf{2.0} & 1.2 & \textbf{100.0} & 14.0 & 11.6 \\
& FFN-2  & 10.2 & 0.5 & 26.2 & \textbf{100.0} & 26.5 & 11.6 \\
& FFN-3  & 17.3 & 0.5 & \textbf{26.8} & \textbf{100.0} & 10.5 & 7.0 \\
& ATTN-1 & 14.2 & 0.0 & 0.6 & \textbf{100.0} & 19.5 & 14.0 \\
& ATTN-2 & 17.8 & 0.0 & 2.4 & \textbf{100.0} & 31.0 & 14.0 \\
& ATTN-3 & \textbf{23.9} & 0.0 & 2.4 & \textbf{100.0} & \textbf{35.5} & \textbf{16.3} \\
\midrule
\multirow{6}{*}{\ace}
& FFN-1  & 35.5 & 95.4 & 13.7 & \textbf{100.0} & 94.5 & 95.3 \\
& FFN-2  & 47.2 & \textbf{99.5} & 10.7 & \textbf{100.0} & 96.5 & 93.0 \\
& FFN-3  & 52.8 & \textbf{99.5} & 13.7 & \textbf{100.0} & \textbf{97.0} & 90.7 \\
& ATTN-1 & 55.3 & 76.6 & 59.5 & \textbf{100.0} & 91.0 & \textbf{97.7} \\
& ATTN-2 & \textbf{63.5} & 98.5 & 67.3 & \textbf{100.0} & 96.0 & \textbf{97.7} \\
& ATTN-3 & 62.9 & 97.5 & \textbf{68.5} & \textbf{100.0} & 96.5 & \textbf{97.7} \\
\bottomrule
\end{tabular}
\end{table*}

\subsection{Choice of Fast Weights}

We compared FFN and attention updates on Qwen3.5-4B and Qwen3.5-9B in Tables~\ref{tab:fast-weights-4b}-\ref{tab:fast-weights-9b}, also examining updates to multiple layers beyond the default final-FFN setting. Here, FFN-$k$ and ATTN-$k$ denote updates to the FFN and attention sublayers in the final $k$ transformer layers, respectively, for $k\in\{1,2,3\}$. For \ace, updating more FFN layers improved Recognition but had mixed effects on other tasks. ATTN-2 and ATTN-3 outperformed all FFN configurations on Recall and Continuation across both models. In particular, ATTN-3 increased Continuation accuracy from $48.8\%$ to $72.0\%$ on 4B and from $13.7\%$ to $68.5\%$ on 9B, relative to FFN-1. TCE benefited from different choices: attention updates yielded its highest Recall accuracy but zero Recognition accuracy, whereas FFN-3 performed best on Continuation. Retention remained at $100\%$ across configurations on 9B but varied on 4B, where \ace\ was substantially more stable than TCE. 
These results highlight the benefits of exploring fast-weight configurations beyond FFN-1, with the best choice depending on the model, method, and task.
A systematic investigation of these choices lies beyond the scope of this work and remains a promising direction for future research.

\clearpage
\section{More Results on Sandbox Tasks}
\label{app:complete-sandbox}

We provide additional sandbox results examining the effects
of encoding thresholds, update budgets, invocation prompts,
UID design, and encoding wording. Selected analyses also
include RULER NIAH-2 as a complementary retrieval task.

\subsection{Sensitivity to the Encoding Threshold}
\label{app:sandbox-threshold-sensitivity}

Tables~\ref{tab:sandbox-threshold-use-4b}--\ref{tab:sandbox-threshold-conflict-27b} report the complete sandbox results across stopping thresholds, extending the main comparisons at $\tau=0.1$. Increasing $\tau$ generally reduced Recall and Recognition accuracy, particularly the latter, whereas its effect on Continuation varied across models and prompts. In contrast, two-sided Counterfactual accuracy tended to improve at larger $\tau$.
Continuation accuracy measures whether the model correctly answered the encoded question. Table~\ref{tab:sandbox-continuation-recovery} additionally reports question-recovery rates, providing a lower bound on retrieval without requiring the model to answer the recovered question. Table~\ref{tab:sandbox-conflict-controls} reports accuracy on non-conflicting control queries paired with Dispute and Counterfactual instances. These controls assess performance on information that should remain unaffected by encoding and are excluded from the primary Conflict scores.

\begin{table*}[t]
\caption{Comparison between TCE and \ace\ on Recall, Recognition,
Continuation, and Retention across stopping thresholds.
We reported accuracy (\%) on Qwen3.5-4B.
Bold marks the best result within each threshold and task.}
\label{tab:sandbox-threshold-use-4b}
\centering
\small
\setlength{\tabcolsep}{5.5pt}
\begin{tabular}{ccccccc}
\toprule
$\tau$ & Method & Prompt & Recall & Recognition & Continuation & Retention \\
\midrule
\multirow{4}{*}{0.1}
& TCE  & --                 & 6.1 & 0.5 & 0.0 & \textbf{100.0} \\
& \ace & \texttt{ci\_short} & 52.3 & \textbf{91.9} & 46.4 & 99.0 \\
& \ace & \texttt{ci\_v1}    & 55.3 & 81.7 & \textbf{48.8} & \textbf{100.0} \\
& \ace & \texttt{ci\_v2}    & \textbf{60.4} & 80.7 & 28.0 & \textbf{100.0} \\
\midrule
\multirow{4}{*}{0.2}
& TCE  & --                 & 6.6 & 0.0 & 0.0 & \textbf{100.0} \\
& \ace & \texttt{ci\_short} & 51.3 & \textbf{86.3} & 45.8 & 99.0 \\
& \ace & \texttt{ci\_v1}    & 51.8 & 72.6 & \textbf{48.2} & \textbf{100.0} \\
& \ace & \texttt{ci\_v2}    & \textbf{57.9} & 71.6 & 28.6 & \textbf{100.0} \\
\midrule
\multirow{4}{*}{1.0}
& TCE  & --                 & 4.1 & 0.0 & 0.0 & \textbf{100.0} \\
& \ace & \texttt{ci\_short} & 38.1 & \textbf{41.6} & 38.1 & 98.0 \\
& \ace & \texttt{ci\_v1}    & 39.1 & 16.2 & \textbf{47.6} & \textbf{100.0} \\
& \ace & \texttt{ci\_v2}    & \textbf{42.1} & 11.2 & 26.8 & \textbf{100.0} \\
\bottomrule
\end{tabular}
\end{table*}

\begin{table*}[t]
\caption{Comparison between TCE and \ace\ on the three categories
of Conflict across stopping thresholds. We reported accuracy (\%)
on Qwen3.5-4B, together with two-sided Counterfactual accuracy.
Bold marks the best result within
each threshold and metric.}
\label{tab:sandbox-threshold-conflict-4b}
\centering
\small
\setlength{\tabcolsep}{5.5pt}
\begin{tabular}{ccccccc}
\toprule
\multirow{2}{*}{$\tau$}
& \multirow{2}{*}{Method}
& \multirow{2}{*}{Prompt}
& \multirow{2}{*}{Counterfact}
& \multirow{2}{*}{Dispute}
& \multirow{2}{*}{Outdated}
& Counterfact \\
& & & & & & (two-sided) \\
\midrule
\multirow{4}{*}{0.1}
& TCE  & --                & 3.5 & 7.0 & 5.9 & 0.5 \\
& \ace & \texttt{ci\_short} & 86.5 & 83.7 & 88.2 & 20.0 \\
& \ace & \texttt{ci\_v1}    & 92.5 & \textbf{95.3} & \textbf{100.0} & 27.5 \\
& \ace & \texttt{ci\_v2}    & \textbf{93.0} & 90.7 & 94.1 & \textbf{28.0} \\
\midrule
\multirow{4}{*}{0.2}
& TCE  & --                & 4.5 & 4.7 & 5.9 & 1.5 \\
& \ace & \texttt{ci\_short} & 86.0 & 79.1 & 88.2 & 21.5 \\
& \ace & \texttt{ci\_v1}    & \textbf{92.5} & \textbf{95.3} & \textbf{100.0} & \textbf{29.0} \\
& \ace & \texttt{ci\_v2}    & \textbf{92.5} & 90.7 & 94.1 & 28.5 \\
\midrule
\multirow{4}{*}{1.0}
& TCE  & --                & 13.5 & 4.7 & 23.5 & 7.0 \\
& \ace & \texttt{ci\_short} & 85.5 & 69.8 & 88.2 & 31.0 \\
& \ace & \texttt{ci\_v1}    & \textbf{93.5} & \textbf{90.7} & \textbf{100.0} & 41.0 \\
& \ace & \texttt{ci\_v2}    & 92.5 & 86.0 & 94.1 & \textbf{41.5} \\
\bottomrule
\end{tabular}
\end{table*}

\begin{table*}[t]
\caption{Comparison between TCE and \ace\ on Recall, Recognition,
Continuation, and Retention across stopping thresholds.
We reported accuracy (\%) on Qwen3.5-9B.
Bold marks the best result within each threshold and task.}
\label{tab:sandbox-threshold-use-9b}
\centering
\small
\setlength{\tabcolsep}{5.5pt}
\begin{tabular}{ccccccc}
\toprule
$\tau$ & Method & Prompt & Recall & Recognition & Continuation & Retention \\
\midrule
\multirow{4}{*}{0.1}
& TCE  & --                 & 4.6 & 2.0 & 1.2 & \textbf{100.0} \\
& \ace & \texttt{ci\_short} & \textbf{40.1} & \textbf{96.4} & 9.5 & \textbf{100.0} \\
& \ace & \texttt{ci\_v1}    & 35.5 & 95.4 & \textbf{13.7} & \textbf{100.0} \\
& \ace & \texttt{ci\_v2}    & 38.1 & 94.4 & 5.4 & \textbf{100.0} \\
\midrule
\multirow{4}{*}{0.2}
& TCE  & --                 & 4.1 & 0.5 & 0.6 & \textbf{100.0} \\
& \ace & \texttt{ci\_short} & \textbf{37.1} & \textbf{86.8} & 8.3 & \textbf{100.0} \\
& \ace & \texttt{ci\_v1}    & 32.0 & 86.3 & \textbf{11.3} & \textbf{100.0} \\
& \ace & \texttt{ci\_v2}    & 36.5 & 82.2 & 4.2 & \textbf{100.0} \\
\midrule
\multirow{4}{*}{1.0}
& TCE  & --                 & 3.6 & 0.0 & 0.0 & \textbf{100.0} \\
& \ace & \texttt{ci\_short} & 26.9 & \textbf{56.9} & 1.2 & \textbf{100.0} \\
& \ace & \texttt{ci\_v1}    & 25.9 & 42.1 & \textbf{6.0} & \textbf{100.0} \\
& \ace & \texttt{ci\_v2}    & \textbf{29.4} & 35.0 & 1.2 & \textbf{100.0} \\
\bottomrule
\end{tabular}
\end{table*}

\begin{table*}[t]
\caption{Comparison between TCE and \ace\ on the three categories
of Conflict across stopping thresholds. We reported accuracy (\%)
on Qwen3.5-9B, together with two-sided Counterfactual accuracy.
Bold marks the best result within
each threshold and metric.}
\label{tab:sandbox-threshold-conflict-9b}
\centering
\small
\setlength{\tabcolsep}{5.5pt}
\begin{tabular}{ccccccc}
\toprule
\multirow{2}{*}{$\tau$}
& \multirow{2}{*}{Method}
& \multirow{2}{*}{Prompt}
& \multirow{2}{*}{Counterfact}
& \multirow{2}{*}{Dispute}
& \multirow{2}{*}{Outdated}
& Counterfact \\
& & & & & & (two-sided) \\
\midrule
\multirow{4}{*}{0.1}
& TCE  & --                 & 14.0 & 11.6 & 10.0 & 6.0 \\
& \ace & \texttt{ci\_short} & 94.0 & 95.3 & \textbf{100.0} & 29.0 \\
& \ace & \texttt{ci\_v1}    & 94.5 & 95.3 & \textbf{100.0} & \textbf{30.5} \\
& \ace & \texttt{ci\_v2}    & \textbf{95.0} & \textbf{97.7} & \textbf{100.0} & \textbf{30.5} \\
\midrule
\multirow{4}{*}{0.2}
& TCE  & --                 & 15.0 & 11.6 & 10.0 & 10.0 \\
& \ace & \texttt{ci\_short} & 90.5 & \textbf{95.3} & \textbf{100.0} & 37.0 \\
& \ace & \texttt{ci\_v1}    & 91.0 & 93.0 & 95.0 & 37.0 \\
& \ace & \texttt{ci\_v2}    & \textbf{94.5} & 93.0 & \textbf{100.0} & \textbf{40.0} \\
\midrule
\multirow{4}{*}{1.0}
& TCE  & --                 & 21.0 & 11.6 & 10.0 & 16.0 \\
& \ace & \texttt{ci\_short} & 76.5 & \textbf{90.7} & \textbf{100.0} & 45.5 \\
& \ace & \texttt{ci\_v1}    & 86.0 & 88.4 & 95.0 & 50.0 \\
& \ace & \texttt{ci\_v2}    & \textbf{91.0} & \textbf{90.7} & \textbf{100.0} & \textbf{54.0} \\
\bottomrule
\end{tabular}
\end{table*}

\begin{table*}[t]
\caption{Comparison between TCE and \ace\ on Recall, Recognition,
Continuation, and Retention across stopping thresholds.
We reported accuracy (\%) on Qwen3.5-27B.
Bold marks the best result within each threshold and task.}
\label{tab:sandbox-threshold-use-27b}
\centering
\small
\setlength{\tabcolsep}{5.5pt}
\begin{tabular}{ccccccc}
\toprule
$\tau$ & Method & Prompt & Recall & Recognition & Continuation & Retention \\
\midrule
\multirow{4}{*}{0.1}
& TCE  & --                 & 1.5 & 0.0 & 0.0 & \textbf{100.0} \\
& \ace & \texttt{ci\_short} & \textbf{75.6} & \textbf{98.0} & \textbf{20.8} & \textbf{100.0} \\
& \ace & \texttt{ci\_v1}    & 39.1 & 58.4 & 11.9 & \textbf{100.0} \\
& \ace & \texttt{ci\_v2}    & 45.2 & 50.8 & 2.4 & \textbf{100.0} \\
\midrule
\multirow{4}{*}{0.2}
& TCE  & --                 & 2.0 & 0.0 & 0.0 & \textbf{100.0} \\
& \ace & \texttt{ci\_short} & \textbf{71.6} & \textbf{83.8} & 0.6 & \textbf{100.0} \\
& \ace & \texttt{ci\_v1}    & 37.1 & 39.6 & \textbf{29.8} & \textbf{100.0} \\
& \ace & \texttt{ci\_v2}    & 44.2 & 28.9 & 1.2 & \textbf{100.0} \\
\midrule
\multirow{4}{*}{1.0}
& TCE  & --                 & 1.5 & 0.0 & 0.0 & \textbf{100.0} \\
& \ace & \texttt{ci\_short} & \textbf{53.3} & \textbf{22.3} & 0.6 & \textbf{100.0} \\
& \ace & \texttt{ci\_v1}    & 28.9 & 1.5 & \textbf{32.1} & \textbf{100.0} \\
& \ace & \texttt{ci\_v2}    & 33.0 & 1.0 & 2.4 & \textbf{100.0} \\
\bottomrule
\end{tabular}
\end{table*}

\begin{table*}[t]
\caption{Comparison between TCE and \ace\ on the three categories
of Conflict across stopping thresholds. We reported accuracy (\%)
on Qwen3.5-27B, together with two-sided Counterfactual accuracy.
 Bold marks the best result within
each threshold and metric.}
\label{tab:sandbox-threshold-conflict-27b}
\centering
\small
\setlength{\tabcolsep}{5.5pt}
\begin{tabular}{ccccccc}
\toprule
\multirow{2}{*}{$\tau$}
& \multirow{2}{*}{Method}
& \multirow{2}{*}{Prompt}
& \multirow{2}{*}{Counterfact}
& \multirow{2}{*}{Dispute}
& \multirow{2}{*}{Outdated}
& Counterfact \\
& & & & & & (two-sided) \\
\midrule
\multirow{4}{*}{0.1}
& TCE  & --                & 11.0 & 18.6 & 54.5 & 8.0 \\
& \ace & \texttt{ci\_short} & \textbf{96.0} & \textbf{97.7} & 97.0 & \textbf{36.5} \\
& \ace & \texttt{ci\_v1}    & 93.0 & 88.4 & \textbf{100.0} & 29.0 \\
& \ace & \texttt{ci\_v2}    & 92.0 & 93.0 & 97.0 & 31.5 \\
\midrule
\multirow{4}{*}{0.2}
& TCE  & --                & 41.0 & 18.6 & 54.5 & 36.0 \\
& \ace & \texttt{ci\_short} & \textbf{96.0} & \textbf{97.7} & \textbf{97.0} & \textbf{38.5} \\
& \ace & \texttt{ci\_v1}    & 92.5 & 86.0 & \textbf{97.0} & 30.5 \\
& \ace & \texttt{ci\_v2}    & 92.0 & 90.7 & 93.9 & 32.5 \\
\midrule
\multirow{4}{*}{1.0}
& TCE  & --                & 43.0 & 9.3 & 24.2 & 37.5 \\
& \ace & \texttt{ci\_short} & \textbf{91.0} & \textbf{93.0} & 90.9 & \textbf{55.0} \\
& \ace & \texttt{ci\_v1}    & 81.5 & 69.8 & \textbf{97.0} & 43.5 \\
& \ace & \texttt{ci\_v2}    & 78.0 & 67.4 & 93.9 & 42.5 \\
\bottomrule
\end{tabular}
\end{table*}

\begin{table*}[t]
\caption{Comparison between TCE and \ace\ on question recovery
in Continuation across stopping thresholds.
We reported question-recovery rates (\%) on the Qwen3.5 family.
These scores provide a lower bound on retrieval, rather than
measuring whether the encoded question was answered correctly.
Bold marks the best result within each model and threshold.}
\label{tab:sandbox-continuation-recovery}
\centering
\small
\setlength{\tabcolsep}{5.5pt}
\begin{tabular}{cccccc}
\toprule
LLM & Method & Prompt & $\tau=0.1$ & $\tau=0.2$ & $\tau=1.0$ \\
\midrule
\multirow{4}{*}{Qwen3.5-4B}
& TCE  & --                 & 0.0 & 0.0 & 0.0 \\
& \ace & \texttt{ci\_short} & 50.0 & 48.2 & 38.1 \\
& \ace & \texttt{ci\_v1}    & \textbf{53.6} & \textbf{52.4} & \textbf{49.4} \\
& \ace & \texttt{ci\_v2}    & 31.5 & 31.0 & 28.6 \\
\midrule
\multirow{4}{*}{Qwen3.5-9B}
& TCE  & --                 & 1.2 & 0.6 & 0.0 \\
& \ace & \texttt{ci\_short} & 28.0 & \textbf{21.4} & 1.2 \\
& \ace & \texttt{ci\_v1}    & \textbf{32.7} & \textbf{21.4} & \textbf{6.0} \\
& \ace & \texttt{ci\_v2}    & 16.7 & 8.3 & 0.6 \\
\midrule
\multirow{4}{*}{Qwen3.5-27B}
& TCE  & --                 & 0.0 & 0.0 & 0.0 \\
& \ace & \texttt{ci\_short} & \textbf{43.5} & 7.1 & 1.8 \\
& \ace & \texttt{ci\_v1}    & 35.1 & \textbf{42.9} & \textbf{37.5} \\
& \ace & \texttt{ci\_v2}    & 20.8 & 9.5 & 5.4 \\
\bottomrule
\end{tabular}
\end{table*}

\begin{table*}[t]
\caption{{Comparison between TCE and \ace\ on non-conflicting
control queries paired with Dispute and Counterfactual instances.
We reported accuracy (\%) on the Qwen3.5 family across stopping
thresholds. Bold marks the best result within each model,
threshold, and control category.}}
\label{tab:sandbox-conflict-controls}
\centering
\small
\setlength{\tabcolsep}{5.5pt}
\begin{tabular}{ccccccccc}
\toprule
\multirow{2}{*}{LLM}
& \multirow{2}{*}{Method}
& \multirow{2}{*}{Prompt}
& \multicolumn{3}{c}{Dispute control}
& \multicolumn{3}{c}{Counterfact control} \\
\cmidrule(lr){4-6}
\cmidrule(lr){7-9}
& & & $\tau=0.1$ & $\tau=0.2$ & $\tau=1.0$
& $\tau=0.1$ & $\tau=0.2$ & $\tau=1.0$ \\
\midrule
\multirow{4}{*}{Qwen3.5-4B}
& TCE  & --
& 74.4 & 79.1 & \textbf{81.4}
& 64.0 & 68.0 & 81.0 \\
& \ace & \texttt{ci\_short}
& 79.1 & \textbf{81.4} & 79.1
& 76.5 & 77.0 & 80.0 \\
& \ace & \texttt{ci\_v1}
& \textbf{81.4} & \textbf{81.4} & 79.1
& 76.0 & 77.0 & 81.5 \\
& \ace & \texttt{ci\_v2}
& \textbf{81.4} & \textbf{81.4} & \textbf{81.4}
& \textbf{79.0} & \textbf{80.0} & \textbf{82.0} \\
\midrule
\multirow{4}{*}{Qwen3.5-9B}
& TCE  & --
& \textbf{81.4} & \textbf{81.4} & \textbf{81.4}
& \textbf{82.5} & \textbf{82.5} & \textbf{82.5} \\
& \ace & \texttt{ci\_short}
& 79.1 & 79.1 & 79.1
& 79.5 & 80.0 & 81.5 \\
& \ace & \texttt{ci\_v1}
& 79.1 & 79.1 & 79.1
& 80.5 & 80.5 & \textbf{82.5} \\
& \ace & \texttt{ci\_v2}
& 76.7 & 79.1 & 79.1
& 79.0 & 80.0 & 81.5 \\
\midrule
\multirow{4}{*}{Qwen3.5-27B}
& TCE  & --
& \textbf{81.4} & \textbf{81.4} & \textbf{81.4}
& \textbf{82.5} & \textbf{82.5} & \textbf{82.5} \\
& \ace & \texttt{ci\_short}
& \textbf{81.4} & \textbf{81.4} & \textbf{81.4}
& 82.0 & 82.0 & \textbf{82.5} \\
& \ace & \texttt{ci\_v1}
& \textbf{81.4} & \textbf{81.4} & \textbf{81.4}
& 82.0 & \textbf{82.5} & \textbf{82.5} \\
& \ace & \texttt{ci\_v2}
& \textbf{81.4} & \textbf{81.4} & \textbf{81.4}
& \textbf{82.5} & \textbf{82.5} & \textbf{82.5} \\
\bottomrule
\end{tabular}
\end{table*}

\subsection{Comparison at Fixed Update Budgets}
\label{sec:fixed-budget}

We compared TCE and \ace\ with exactly
$K\in\{2,4,8,16,32\}$ updates per chunk, replacing adaptive
stopping while retaining their encoding objectives and
inference procedures, with \ace\ using \texttt{ci\_v1}.
We used the same $197$ sandbox instances for Recall and
Recognition and $100$ RULER NIAH-2 instances on
Qwen3.5-4B, 9B, and 27B.
This matched update counts rather than computational costs,
as each \ace\ update used two copies of the context.
Tables~\ref{tab:fixed-budget-4b}--\ref{tab:fixed-budget-27b}
show that \ace\ achieved higher accuracy in $35$ of $36$
comparisons at $K\geq4$, with one tie.
Recognition reached $98.5$--$99.5\%$ by $K=8$ with \ace.
Although additional updates improved TCE on Recognition
and generally improved retrieval on NIAH-2, TCE remained
below \ace\ at larger budgets.
These results suggest that the gains of \ace\ were not
solely due to additional encoding updates.

\begin{table*}[t]
\caption{Comparison between TCE and \ace\ across fixed
update budgets on Qwen3.5-4B.
We reported accuracy (\%) on Recall, Recognition,
and RULER NIAH-2, with \ace\ using \texttt{ci\_v1}.
Bold marks the best result within each budget and task.}
\label{tab:fixed-budget-4b}
\centering
\small
\setlength{\tabcolsep}{5.5pt}
\begin{tabular}{ccccc}
\toprule
Updates $K$ & Method & Recall & Recognition & NIAH-2 \\
\midrule
\multirow{2}{*}{2}
& TCE  & 3.0 & \textbf{0.0} & \textbf{0.0} \\
& \ace & \textbf{15.2} & \textbf{0.0} & \textbf{0.0} \\
\midrule
\multirow{2}{*}{4}
& TCE  & 1.5 & 1.5 & 0.0 \\
& \ace & \textbf{45.7} & \textbf{62.4} & \textbf{1.0} \\
\midrule
\multirow{2}{*}{8}
& TCE  & 2.0 & 27.9 & 8.0 \\
& \ace & \textbf{58.9} & \textbf{98.5} & \textbf{16.0} \\
\midrule
\multirow{2}{*}{16}
& TCE  & 1.0 & 69.0 & 9.0 \\
& \ace & \textbf{59.4} & \textbf{98.5} & \textbf{37.0} \\
\midrule
\multirow{2}{*}{32}
& TCE  & 1.0 & 76.1 & 14.0 \\
& \ace & \textbf{57.4} & \textbf{98.5} & \textbf{46.0} \\
\bottomrule
\end{tabular}
\end{table*}

\begin{table*}[t]
\caption{Comparison between TCE and \ace\ across fixed
update budgets on Qwen3.5-9B.
We reported accuracy (\%) on Recall, Recognition,
and RULER NIAH-2, with \ace\ using \texttt{ci\_v1}.
Bold marks the best result within each budget and task.}
\label{tab:fixed-budget-9b}
\centering
\small
\setlength{\tabcolsep}{5.5pt}
\begin{tabular}{ccccc}
\toprule
Updates $K$ & Method & Recall & Recognition & NIAH-2 \\
\midrule
\multirow{2}{*}{2}
& TCE  & 3.0 & \textbf{0.0} & \textbf{0.0} \\
& \ace & \textbf{10.2} & \textbf{0.0} & \textbf{0.0} \\
\midrule
\multirow{2}{*}{4}
& TCE  & 4.1 & 1.0 & 0.0 \\
& \ace & \textbf{34.0} & \textbf{79.2} & \textbf{1.0} \\
\midrule
\multirow{2}{*}{8}
& TCE  & 0.5 & 70.6 & 5.0 \\
& \ace & \textbf{44.7} & \textbf{99.5} & \textbf{8.0} \\
\midrule
\multirow{2}{*}{16}
& TCE  & 2.0 & 76.6 & 9.0 \\
& \ace & \textbf{50.3} & \textbf{99.5} & \textbf{14.0} \\
\midrule
\multirow{2}{*}{32}
& TCE  & 2.5 & 75.1 & 8.0 \\
& \ace & \textbf{50.8} & \textbf{99.5} & \textbf{19.0} \\
\bottomrule
\end{tabular}
\end{table*}

\begin{table*}[t]
\caption{Comparison between TCE and \ace\ across fixed
update budgets on Qwen3.5-27B.
We reported accuracy (\%) on Recall, Recognition,
and RULER NIAH-2, with \ace\ using \texttt{ci\_v1}.
Bold marks the best result within each budget and task.}
\label{tab:fixed-budget-27b}
\centering
\small
\setlength{\tabcolsep}{5.5pt}
\begin{tabular}{ccccc}
\toprule
Updates $K$ & Method & Recall & Recognition & NIAH-2 \\
\midrule
\multirow{2}{*}{2}
& TCE  & 0.5 & \textbf{0.0} & \textbf{0.0} \\
& \ace & \textbf{12.7} & \textbf{0.0} & \textbf{0.0} \\
\midrule
\multirow{2}{*}{4}
& TCE  & 2.0 & 0.0 & \textbf{0.0} \\
& \ace & \textbf{40.1} & \textbf{15.2} & \textbf{0.0} \\
\midrule
\multirow{2}{*}{8}
& TCE  & 2.0 & 6.6 & 5.0 \\
& \ace & \textbf{42.6} & \textbf{99.0} & \textbf{21.0} \\
\midrule
\multirow{2}{*}{16}
& TCE  & 2.5 & 39.6 & 8.0 \\
& \ace & \textbf{44.7} & \textbf{99.0} & \textbf{42.0} \\
\midrule
\multirow{2}{*}{32}
& TCE  & 3.6 & 44.7 & 12.0 \\
& \ace & \textbf{44.2} & \textbf{99.0} & \textbf{46.0} \\
\bottomrule
\end{tabular}
\end{table*}
\subsection{Effect of Invocation on Prior-Knowledge Recovery}
\label{app:invocation-prior}

We examined whether the context-invocation prompt
affected the ability of \ace\ to set aside encoded information
and answer from pre-existing knowledge.
Using the same $200$ Counterfactual instances as in the main
sandbox experiments, we encoded each context with $\tau=0.1$
and evaluated the prior-knowledge query with and without
\texttt{ci\_short}, keeping the query and encoded model weights
unchanged.
Table~\ref{tab:invocation-prior} reports the results alongside
contextual-answer accuracy for TCE and \ace.
While \ace\ substantially improved contextual-answer accuracy,
removing the invocation had mixed effects on prior-knowledge
recovery. Accuracy increased by $6.0$ percentage points on
Qwen3.5-4B but decreased by $4.5$ and $3.5$ points on
Qwen3.5-9B and Qwen3.5-27B, respectively.
Thus, removing the invocation alone did not consistently
improve prior-knowledge recovery, suggesting that the
difficulty was not solely due to the invocation prompt.

\begin{table*}[t]
\caption{Comparison between TCE and \ace\ on contextual
answering and the effect of invocation on prior-knowledge
recovery. We reported accuracy (\%) on $200$ Counterfactual
instances per model with $\tau=0.1$.
TCE used no invocation, while \ace\ used \texttt{ci\_short}. The two prior-answer conditions
used the same encoded model state and query.}
\label{tab:invocation-prior}
\centering
\small
\setlength{\tabcolsep}{5.5pt}
\begin{tabular}{ccccc}
\toprule
\multirow{2}{*}{LLM}
& \multicolumn{2}{c}{Contextual answer}
& \multicolumn{2}{c}{Prior answer (\ace)} \\
\cmidrule(lr){2-3}\cmidrule(lr){4-5}
& TCE & \ace & With invocation & Without invocation \\
\midrule
Qwen3.5-4B  & 3.5  & 86.5 & 25.5 & 31.5 \\
Qwen3.5-9B  & 14.0 & 94.0 & 33.5 & 29.0 \\
Qwen3.5-27B & 11.0 & 96.0 & 39.0 & 35.5 \\
\bottomrule
\end{tabular}
\end{table*}

\subsection{TCE with Generic Invocation}
\label{sec:generic-invocation}

We examined whether adding an invocation prompt
to TCE, without changing its encoding procedure, could
recover the gains of \ace.
We retained the original autoregressive encoding objective
and prepended the following generic invocation to the task query:
\begin{diagnosticbox}[Generic Invocation]
\textit{Recall the previously encoded context and use information
relevant to the following request.}
\end{diagnosticbox}
This prompt instructs the model to use encoded information
without specifying UIDs or timestamps.
We compared TCE with and without this invocation against
full \ace\ on the same $197$ sandbox instances for Recall
and Recognition, with an additional evaluation on $100$
RULER NIAH-2 instances.
We used Qwen3.5-4B, 9B, and 27B with $\tau=0.1$.
The original TCE and \ace\ results served as references,
with \ace\ using \texttt{ci\_v1} on the sandbox tasks
and \texttt{ci\_v2} on NIAH-2.
Table~\ref{tab:generic-invocation} shows that adding the
generic invocation had limited and inconsistent benefits.
Although it improved Recall on Qwen3.5-9B and Qwen3.5-27B,
Recognition accuracy remained at $0.0$--$1.5\%$, and
NIAH-2 accuracy ranged from $5.0\%$ to $10.0\%$.
Performance remained substantially below full \ace\
across all evaluated models and tasks.
These results suggest that adding a recall instruction
at inference alone was insufficient to overcome the
limitations of autoregressive TCE in these settings.

\begin{table*}[t]
\caption{Comparison between TCE, TCE with generic invocation,
and \ace\ on Recall, Recognition, and RULER NIAH-2.
We reported accuracy (\%) on the Qwen3.5 family with $\tau=0.1$.
Bold marks the best result within each model and task.}
\label{tab:generic-invocation}
\centering
\small
\setlength{\tabcolsep}{5.5pt}
\begin{tabular}{ccccc}
\toprule
LLM & Method & Recall & Recognition & NIAH-2 \\
\midrule
\multirow{3}{*}{Qwen3.5-4B}
& TCE & 6.1 & 0.5 & 7.0 \\
& TCE with generic invocation & 2.0 & 0.0 & 10.0 \\
& \ace & \textbf{55.3} & \textbf{81.7} & \textbf{35.0} \\
\midrule
\multirow{3}{*}{Qwen3.5-9B}
& TCE & 4.6 & 2.0 & 7.0 \\
& TCE with generic invocation & 9.1 & 1.5 & 7.0 \\
& \ace & \textbf{35.5} & \textbf{95.4} & \textbf{17.0} \\
\midrule
\multirow{3}{*}{Qwen3.5-27B}
& TCE & 1.5 & 0.0 & 7.0 \\
& TCE with generic invocation & 15.7 & 0.0 & 5.0 \\
& \ace & \textbf{39.1} & \textbf{58.4} & \textbf{40.0} \\
\bottomrule
\end{tabular}
\end{table*}

\subsection{Unique versus Shared UIDs}
\label{sec:common-marker}

We examined whether \ace\ required unique UIDs or could
instead use the same UID across contexts.
We replaced every UID occurrence with the fixed string
\texttt{memory} in both encoding sequences and the
\texttt{ci\_v1} invocation, while retaining the original
timestamps and all other settings.
For terminological consistency, we refer to this shared
identifier as a UID, although it is no longer unique
across contexts.
Each instance was encoded independently and contained
a single context chunk.
We evaluated Recognition on all $197$ sandbox instances
using Qwen3.5-4B, 9B, and 27B at $\tau=0.1$.
Table~\ref{tab:common-marker} shows that using the shared
UID reduced Recognition accuracy across all three models.
These results favor the original UID design for Recognition
in this single-context setting.

\begin{table}[t]
\caption{Comparison between \ace\ with a shared UID and
the original \ace\ on Recognition.
We reported accuracy (\%) on the Qwen3.5 family with
$\tau=0.1$.
The shared UID was the fixed string \texttt{memory}.
Bold marks the better result within each model.}
\label{tab:common-marker}
\centering
\small
\setlength{\tabcolsep}{5.5pt}
\begin{tabular}{ccc}
\toprule
LLM & \ace\ with shared UID & \ace \\
\midrule
Qwen3.5-4B  & 66.0 & \textbf{81.7} \\
Qwen3.5-9B  & 91.9 & \textbf{95.4} \\
Qwen3.5-27B & 47.7 & \textbf{58.4} \\
\bottomrule
\end{tabular}
\end{table}

\subsection{Effect of Encoding Wording}
\label{sec:neutral-prefix}

We further examined the effect of the role-binding prompt
by replacing it with a more neutral formulation.
The alternative retained the UID and timestamp but did
not explicitly instruct the model to treat the encoded
information as conversational context:
\begin{diagnosticbox}[Alternative Role-Binding Prompt]
\textit{The following text is accompanied by a timestamp with
the specified value $s_t$, identified by $i_t$.
The identifier simply serves as a label for this text,
and the text associated with $i_t$ is below: $c_t$}
\end{diagnosticbox}
The alternative prefix matched the original in token length
and preserved the occurrence counts of the UID and timestamp.
We kept the UID-conditioned encoding objective, input format,
supervised token positions, and loss aggregation unchanged.
Both variants therefore used two copies of the context
per update.
We evaluated the same $197$ sandbox instances for Recall
and Recognition, with an additional evaluation on $100$
RULER NIAH-2 instances, using Qwen3.5-4B, 9B, and 27B
with $\tau=0.1$ and \texttt{ci\_v1}.
Table~\ref{tab:neutral-prefix} shows that the alternative
wording had mixed effects across models and tasks.
Recall accuracy remained similar, while NIAH-2 accuracy
increased on 4B and 27B but decreased slightly on 9B.
Recognition accuracy declined on 4B and 9B but increased
on 27B.
These results suggest that \ace\ supported alternative
encoding formulations, while the preferred wording
depended on the model and task.

\begin{table*}[t]
\caption{Comparison between \ace\ with alternative encoding
wording and the original \ace\ on Recall, Recognition,
and RULER NIAH-2. We reported accuracy (\%) on the Qwen3.5
family with $\tau=0.1$.
Bold marks the best result within each model and task.}
\label{tab:neutral-prefix}
\centering
\small
\setlength{\tabcolsep}{5.5pt}
\begin{tabular}{ccccc}
\toprule
LLM & Method & Recall & Recognition & NIAH-2 \\
\midrule
\multirow{2}{*}{Qwen3.5-4B}
& \ace\ with alternative wording
& 54.3 & 28.9 & \textbf{33.0} \\
& \ace
& \textbf{55.3} & \textbf{81.7} & 32.0 \\
\midrule
\multirow{2}{*}{Qwen3.5-9B}
& \ace\ with alternative wording
& 34.5 & 77.7 & 14.0 \\
& \ace
& \textbf{35.5} & \textbf{95.4} & \textbf{15.0} \\
\midrule
\multirow{2}{*}{Qwen3.5-27B}
& \ace\ with alternative wording
& \textbf{41.6} & \textbf{76.1} & \textbf{47.0} \\
& \ace
& 39.1 & 58.4 & 39.0 \\
\bottomrule
\end{tabular}
\end{table*}

\clearpage
\section{More Results on Long-context Tasks}\label{app:more long context}
Complementing the main evaluations on \textsc{RULER}, \textsc{LongBench}, and \textsc{LongBench v2}, we provide additional analyses of stopping-threshold sensitivity, general capability retention, and the UID addressing cue. We further extend \textsc{RULER} to $4$K and $64$K contexts to assess retrieval after longer sequences of chunk-wise updates.

\subsection{Sensitivity to the Encoding Threshold}
\label{app:threshold-sensitivity}

We examined sensitivity to the stopping threshold $\tau$ on \textsc{RULER} and \textsc{LongBench}, supplementing the main long-context evaluations with additional thresholds. Tables~\ref{tab:ruler-complete-thresholds}--\ref{tab:longbench-v1-complete-performance} report the results.
On \textsc{RULER}, smaller $\tau$ generally improved retrieval, particularly on the harder tasks, although the effect varied across models and prompts. On \textsc{LongBench}, \ace\ outperformed TCE at every reported threshold, but smaller $\tau$ did not consistently yield better results. Overall, the gains from \ace\ persisted across thresholds, while the best threshold depended on the model and task.

\begin{table*}[t]
\caption{Comparison between TCE and \ace\ on RULER across
stopping thresholds for Qwen3.5-4B and Qwen3.5-9B.
We reported retrieval accuracy (\%).
The three \ace\ prompts were applied to the same encoded
state within each model, task, and threshold.
Bold marks the best result within each model, threshold,
and task.}
\label{tab:ruler-complete-thresholds}
\centering
\small
\setlength{\tabcolsep}{5.5pt}
\begin{tabular}{ccccccc}
\toprule
LLM & $\tau$ & Method & Prompt & NIAH-1 & NIAH-2 & NIAH-3 \\
\midrule
\multirow{12}{*}{Qwen3.5-4B}
& \multirow{4}{*}{0.1}
& TCE & -- & 98.0 & 7.0 & 0.0 \\
& & \ace & \texttt{ci\_short} & 99.0 & 30.0 & \textbf{18.0} \\
& & \ace & \texttt{ci\_v1} & \textbf{100.0} & 32.0 & 13.0 \\
& & \ace & \texttt{ci\_v2} & \textbf{100.0} & \textbf{35.0} & 15.0 \\
\cmidrule(lr){2-7}
& \multirow{4}{*}{0.2}
& TCE & -- & \textbf{100.0} & 8.0 & 0.0 \\
& & \ace & \texttt{ci\_short} & 99.0 & 26.0 & \textbf{19.0} \\
& & \ace & \texttt{ci\_v1} & \textbf{100.0} & 30.0 & 12.0 \\
& & \ace & \texttt{ci\_v2} & \textbf{100.0} & \textbf{33.0} & 12.0 \\
\cmidrule(lr){2-7}
& \multirow{4}{*}{1.0}
& TCE & -- & 84.0 & 4.0 & 0.0 \\
& & \ace & \texttt{ci\_short} & \textbf{99.0} & 18.0 & \textbf{9.0} \\
& & \ace & \texttt{ci\_v1} & \textbf{99.0} & \textbf{20.0} & 8.0 \\
& & \ace & \texttt{ci\_v2} & \textbf{99.0} & 19.0 & 8.0 \\
\midrule
\multirow{12}{*}{Qwen3.5-9B}
& \multirow{4}{*}{0.1}
& TCE & -- & 88.0 & 7.0 & 1.0 \\
& & \ace & \texttt{ci\_short} & \textbf{100.0} & 11.0 & \textbf{30.0} \\
& & \ace & \texttt{ci\_v1} & \textbf{100.0} & 15.0 & 25.0 \\
& & \ace & \texttt{ci\_v2} & \textbf{100.0} & \textbf{17.0} & 27.0 \\
\cmidrule(lr){2-7}
& \multirow{4}{*}{0.2}
& TCE & -- & 84.0 & 7.0 & 1.0 \\
& & \ace & \texttt{ci\_short} & \textbf{100.0} & 10.0 & \textbf{28.0} \\
& & \ace & \texttt{ci\_v1} & \textbf{100.0} & 15.0 & 21.0 \\
& & \ace & \texttt{ci\_v2} & \textbf{100.0} & \textbf{16.0} & 22.0 \\
\cmidrule(lr){2-7}
& \multirow{4}{*}{1.0}
& TCE & -- & 57.0 & 3.0 & 1.0 \\
& & \ace & \texttt{ci\_short} & 94.0 & \textbf{8.0} & \textbf{19.0} \\
& & \ace & \texttt{ci\_v1} & \textbf{99.0} & \textbf{8.0} & 15.0 \\
& & \ace & \texttt{ci\_v2} & \textbf{99.0} & \textbf{8.0} & 15.0 \\
\bottomrule
\end{tabular}
\end{table*}

\begin{table*}[t]
\caption{Comparison between TCE and \ace\ on RULER across
stopping thresholds for Qwen3.5-27B and Muse-Glimmer-30B.
We reported retrieval accuracy (\%).
The three \ace\ prompts were applied to the same encoded
state within each model, task, and threshold.
Bold marks the best result within each model, threshold,
and task.}
\label{tab:ruler-complete-thresholds-large}
\centering
\small
\setlength{\tabcolsep}{5.5pt}
\begin{tabular}{ccccccc}
\toprule
LLM & $\tau$ & Method & Prompt & NIAH-1 & NIAH-2 & NIAH-3 \\
\midrule
\multirow{12}{*}{Qwen3.5-27B}
& \multirow{4}{*}{0.1}
& TCE & -- & 85.0 & 7.0 & 0.0 \\
& & \ace & \texttt{ci\_short} & \textbf{100.0} & \textbf{45.0} & \textbf{33.0} \\
& & \ace & \texttt{ci\_v1} & \textbf{100.0} & 39.0 & 31.0 \\
& & \ace & \texttt{ci\_v2} & \textbf{100.0} & 40.0 & 29.0 \\
\cmidrule(lr){2-7}
& \multirow{4}{*}{0.2}
& TCE & -- & 80.0 & 9.0 & 0.0 \\
& & \ace & \texttt{ci\_short} & \textbf{99.0} & \textbf{41.0} & \textbf{33.0} \\
& & \ace & \texttt{ci\_v1} & \textbf{99.0} & 37.0 & 26.0 \\
& & \ace & \texttt{ci\_v2} & \textbf{99.0} & 37.0 & 26.0 \\
\cmidrule(lr){2-7}
& \multirow{4}{*}{1.0}
& TCE & -- & 45.0 & 1.0 & 0.0 \\
& & \ace & \texttt{ci\_short} & \textbf{96.0} & \textbf{17.0} & \textbf{27.0} \\
& & \ace & \texttt{ci\_v1} & 94.0 & 6.0 & 21.0 \\
& & \ace & \texttt{ci\_v2} & 93.0 & 6.0 & 21.0 \\
\midrule
\multirow{12}{*}{Muse-Glimmer-30B}
& \multirow{4}{*}{0.1}
& TCE & -- & 27.0 & 33.0 & 8.0 \\
& & \ace & \texttt{ci\_short} & 79.0 & 29.0 & 54.0 \\
& & \ace & \texttt{ci\_v1} & \textbf{93.0} & 40.0 & \textbf{63.0} \\
& & \ace & \texttt{ci\_v2} & 89.0 & \textbf{43.0} & \textbf{63.0} \\
\cmidrule(lr){2-7}
& \multirow{4}{*}{0.2}
& TCE & -- & 24.0 & 34.0 & 8.0 \\
& & \ace & \texttt{ci\_short} & 79.0 & 34.0 & 54.0 \\
& & \ace & \texttt{ci\_v1} & \textbf{93.0} & 38.0 & 61.0 \\
& & \ace & \texttt{ci\_v2} & 89.0 & \textbf{43.0} & \textbf{63.0} \\
\cmidrule(lr){2-7}
& \multirow{4}{*}{1.0}
& TCE & -- & 5.0 & 20.0 & 8.0 \\
& & \ace & \texttt{ci\_short} & \textbf{88.0} & 34.0 & 57.0 \\
& & \ace & \texttt{ci\_v1} & 76.0 & 33.0 & 56.0 \\
& & \ace & \texttt{ci\_v2} & 82.0 & \textbf{38.0} & \textbf{58.0} \\
\bottomrule
\end{tabular}
\end{table*}

\begin{table*}[t]
\caption{Comparison between TCE and \ace\ on LongBench across
stopping thresholds. We reported answer F1 ($\times100$)
under sequential per-chunk encoding.
Gold Standard refers to full-context inference, which does
not depend on $\tau$.
Bold marks the better result between TCE and \ace\
within each model and threshold.}
\label{tab:longbench-v1-complete-performance}
\centering
\small
\setlength{\tabcolsep}{5.5pt}
\begin{tabular}{ccccc}
\toprule
LLM & Method & $\tau=5$ & $\tau=8$ & $\tau=10$ \\
\midrule
\multirow{3}{*}{Qwen3.5-4B}
& Gold Standard & 20.8 & 20.8 & 20.8 \\
\cmidrule(lr){2-5}
& TCE  & 7.0 & 8.8 & 8.6 \\
& \ace & \textbf{7.8} & \textbf{9.0} & \textbf{10.0} \\
\midrule
\multirow{3}{*}{Qwen3.5-9B}
& Gold Standard & 20.3 & 20.3 & 20.3 \\
\cmidrule(lr){2-5}
& TCE  & 7.4 & 8.2 & 9.2 \\
& \ace & \textbf{10.8} & \textbf{9.6} & \textbf{10.1} \\
\midrule
\multirow{3}{*}{Qwen3.5-27B}
& Gold Standard & 18.9 & 18.9 & 18.9 \\
\cmidrule(lr){2-5}
& TCE  & 8.0 & 7.9 & 8.1 \\
& \ace & \textbf{10.2} & \textbf{9.8} & \textbf{9.6} \\
\midrule
\multirow{3}{*}{Muse-Glimmer-30B}
& Gold Standard & 21.2 & 21.2 & 21.2 \\
\cmidrule(lr){2-5}
& TCE  & 13.0 & 14.0 & 12.9 \\
& \ace & \textbf{13.4} & \textbf{19.2} & \textbf{21.2} \\
\bottomrule
\end{tabular}
\end{table*}

\subsection{General Capability Retention}
\label{app:mmlu-retention}
\label{app:mmlu-ruler}

To assess general ability after TCE and \ace\, we evaluated the encoded LLMs on a fixed subset of $200$ multiple-choice questions from MMLU~\citep{hendrycks2020measuring}, referred to as MMLU-200. These questions are unrelated to the encoded context and provide a supplementary check beyond the context-specific tasks.
Tables~\ref{tab:mmlu-ruler-4b}--\ref{tab:mmlu-ruler-muse} report answer accuracies (\%) across different
stopping thresholds $\tau$, counting unparsable responses as incorrect. TCE was evaluated without an invocation prompt, whereas \ace\ was evaluated with each of the three invocation prompts applied to the same encoded model state, even though the subsequent MMLU questions were unrelated to the encoded context.

For \ace, this evaluation tested robustness to invocation prompts that referenced context unrelated to the current request. The larger LLMs, Qwen3.5-27B and Muse-Glimmer-30B, maintained strong MMLU-200 performance, generally matching or exceeding TCE. The smaller LLMs, Qwen3.5-4B and Qwen3.5-9B, often performed worse than TCE, with more pronounced reductions for Qwen3.5-9B under stricter thresholds and longer invocation prompts. Nevertheless, these models generally retained ability to answer unrelated questions despite the potentially distracting prompts.
These results suggested greater robustness in the larger models. However, without a matched evaluation of the original model, they characterized post-encoding performance rather than the change in general capability caused by encoding.

\begin{table*}[t]
\caption{Comparison between TCE and \ace\ on general capability
retention with Qwen3.5-4B. We reported MMLU-200 accuracy (\%)
after encoding contexts from each RULER task.
Bold marks the best result within each threshold and task.}
\label{tab:mmlu-ruler-4b}
\centering
\small
\setlength{\tabcolsep}{5.5pt}
\begin{tabular}{cccccc}
\toprule
$\tau$ & Method & Prompt & NIAH-1 & NIAH-2 & NIAH-3 \\
\midrule
\multirow{4}{*}{0.1}
& TCE  & --                 & 71.0 & \textbf{76.0} & \textbf{75.3} \\
& \ace & \texttt{ci\_short} & \textbf{72.3} & 75.8 & \textbf{75.3} \\
& \ace & \texttt{ci\_v1}    & 72.0 & 73.8 & 73.3 \\
& \ace & \texttt{ci\_v2}    & 71.0 & 72.5 & 72.8 \\
\midrule
\multirow{4}{*}{0.2}
& TCE  & --                 & 71.5 & \textbf{76.0} & \textbf{75.5} \\
& \ace & \texttt{ci\_short} & \textbf{72.8} & 75.5 & 75.0 \\
& \ace & \texttt{ci\_v1}    & 71.8 & 73.8 & 73.8 \\
& \ace & \texttt{ci\_v2}    & 71.0 & 72.5 & 73.0 \\
\midrule
\multirow{4}{*}{1.0}
& TCE  & --                 & \textbf{73.0} & \textbf{76.3} & \textbf{75.5} \\
& \ace & \texttt{ci\_short} & \textbf{73.0} & 74.8 & 75.3 \\
& \ace & \texttt{ci\_v1}    & 71.5 & 74.8 & 73.5 \\
& \ace & \texttt{ci\_v2}    & 71.5 & 74.0 & 73.5 \\
\bottomrule
\end{tabular}
\end{table*}

\begin{table*}[t]
\caption{Comparison between TCE and \ace\ on general capability
retention with Qwen3.5-9B. We reported MMLU-200 accuracy (\%)
after encoding contexts from each RULER task.
Bold marks the best result within each threshold and task.}
\label{tab:mmlu-ruler-9b}
\centering
\small
\setlength{\tabcolsep}{5.5pt}
\begin{tabular}{cccccc}
\toprule
$\tau$ & Method & Prompt & NIAH-1 & NIAH-2 & NIAH-3 \\
\midrule
\multirow{4}{*}{0.1}
& TCE  & --                 & \textbf{75.5} & \textbf{77.0} & \textbf{76.5} \\
& \ace & \texttt{ci\_short} & 67.0 & 74.5 & 73.5 \\
& \ace & \texttt{ci\_v1}    & 62.5 & 72.5 & 70.0 \\
& \ace & \texttt{ci\_v2}    & 58.8 & 71.0 & 67.5 \\
\midrule
\multirow{4}{*}{0.2}
& TCE  & --                 & \textbf{75.8} & \textbf{77.3} & \textbf{76.8} \\
& \ace & \texttt{ci\_short} & 68.8 & 74.5 & 74.0 \\
& \ace & \texttt{ci\_v1}    & 63.5 & 73.8 & 71.5 \\
& \ace & \texttt{ci\_v2}    & 60.8 & 72.8 & 70.3 \\
\midrule
\multirow{4}{*}{1.0}
& TCE  & --                 & \textbf{75.5} & \textbf{77.8} & \textbf{77.8} \\
& \ace & \texttt{ci\_short} & 73.0 & 76.3 & 75.0 \\
& \ace & \texttt{ci\_v1}    & 71.0 & 73.5 & 73.0 \\
& \ace & \texttt{ci\_v2}    & 69.5 & 72.0 & 72.5 \\
\bottomrule
\end{tabular}
\end{table*}

\begin{table*}[t]
\caption{Comparison between TCE and \ace\ on general capability
retention with Qwen3.5-27B. We reported MMLU-200 accuracy (\%)
after encoding contexts from each RULER task.
Bold marks the best result within each threshold and task.}
\label{tab:mmlu-ruler-27b}
\centering
\small
\setlength{\tabcolsep}{5.5pt}
\begin{tabular}{cccccc}
\toprule
$\tau$ & Method & Prompt & NIAH-1 & NIAH-2 & NIAH-3 \\
\midrule
\multirow{4}{*}{0.1}
& TCE  & --                 & 79.3 & 80.0 & 80.3 \\
& \ace & \texttt{ci\_short} & \textbf{82.0} & \textbf{81.8} & \textbf{81.5} \\
& \ace & \texttt{ci\_v1}    & 80.8 & 80.8 & 80.8 \\
& \ace & \texttt{ci\_v2}    & 81.0 & 80.5 & 79.5 \\
\midrule
\multirow{4}{*}{0.2}
& TCE  & --                 & 79.3 & 80.0 & 80.3 \\
& \ace & \texttt{ci\_short} & \textbf{82.0} & \textbf{81.8} & \textbf{81.3} \\
& \ace & \texttt{ci\_v1}    & 80.5 & 80.8 & 81.0 \\
& \ace & \texttt{ci\_v2}    & 81.3 & 80.8 & 79.5 \\
\midrule
\multirow{4}{*}{1.0}
& TCE  & --                 & 79.8 & 81.0 & 80.0 \\
& \ace & \texttt{ci\_short} & \textbf{82.3} & 81.0 & \textbf{81.3} \\
& \ace & \texttt{ci\_v1}    & 81.0 & \textbf{82.0} & 80.8 \\
& \ace & \texttt{ci\_v2}    & 81.5 & 81.0 & 79.8 \\
\bottomrule
\end{tabular}
\end{table*}

\begin{table*}[t]
\caption{Comparison between TCE and \ace\ on general capability
retention with Muse-Glimmer-30B. We reported MMLU-200 accuracy (\%)
after encoding contexts from each RULER task.
Bold marks the best result within each threshold and task.}
\label{tab:mmlu-ruler-muse}
\centering
\small
\setlength{\tabcolsep}{5.5pt}
\begin{tabular}{cccccc}
\toprule
$\tau$ & Method & Prompt & NIAH-1 & NIAH-2 & NIAH-3 \\
\midrule
\multirow{4}{*}{0.1}
& TCE  & --                 & 66.3 & 72.8 & 69.5 \\
& \ace & \texttt{ci\_short} & 76.5 & 76.0 & 76.0 \\
& \ace & \texttt{ci\_v1}    & 75.5 & 75.0 & 75.3 \\
& \ace & \texttt{ci\_v2}    & \textbf{76.8} & \textbf{76.8} & \textbf{76.5} \\
\midrule
\multirow{4}{*}{0.2}
& TCE  & --                 & 66.3 & 71.3 & 64.3 \\
& \ace & \texttt{ci\_short} & 76.5 & 76.0 & 76.3 \\
& \ace & \texttt{ci\_v1}    & 75.5 & 75.8 & 75.5 \\
& \ace & \texttt{ci\_v2}    & \textbf{76.8} & \textbf{76.8} & \textbf{76.8} \\
\midrule
\multirow{4}{*}{1.0}
& TCE  & --                 & 68.8 & 67.5 & 55.0 \\
& \ace & \texttt{ci\_short} & 76.0 & 75.3 & 74.5 \\
& \ace & \texttt{ci\_v1}    & 75.8 & 74.5 & 74.3 \\
& \ace & \texttt{ci\_v2}    & \textbf{77.0} & \textbf{76.3} & \textbf{75.5} \\
\bottomrule
\end{tabular}
\end{table*}

\begin{table}[t]
\caption{UID ablation for \ace\ on NIAH-1 in \textsc{RULER}. We reported retrieval accuracy (\%) on the Qwen3.5 family. Both conditions use the same encoded state, with only the correct or decoy UID as the addressing cue at inference. Bold marks the better result within each model.}
\label{tab:cid-ablation-ruler}
\centering
\small
\setlength{\tabcolsep}{5.5pt}
\begin{tabular}{ccc}
\toprule
LLM & Correct UID & Decoy UID  \\
\midrule
Qwen3.5-4B  & \textbf{87.0}  & 1.0 \\
Qwen3.5-9B  & \textbf{84.0}  & 8.0 \\
Qwen3.5-27B & \textbf{100.0} & 4.0 \\
\bottomrule
\end{tabular}
\end{table}

\subsection{Effect of the Addressing Cue}
\label{app:addressing-cue}

We further examined whether the UID guides access to encoded information on NIAH-1 in \textsc{RULER}. After full \ace\ encoding, we used only the correct UID or a character-length-matched decoy as the addressing cue, keeping the encoded model fixed. Table~\ref{tab:cid-ablation-ruler} shows that replacing the correct UID reduced retrieval accuracy from $84$--$100$\% to $1$--$8$\%, demonstrating a strong dependence on the addressing cue in this setting.

\subsection{RULER with Longer Contexts}
\label{app:ruler-4k}

We further tested context encoding on \textsc{RULER} with longer contexts, extending NIAH-1 to $64$K tokens and NIAH-2 and NIAH-3 to $4$K tokens. NIAH-1 uses a repetitive background to test retrieval after many sequential updates, whereas NIAH-2 and NIAH-3 require retrieving a seven-digit number and a UUID from essay-based contexts, respectively. We encoded contexts sequentially in approximately $1{,}024$-token chunks, using $\tau=0.1$ and \texttt{ci\_v1} for \ace.
Table~\ref{tab:ruler-64k} shows that on NIAH-1 with $64$K contexts, \ace\ outperformed TCE on all three evaluated models, reaching $80.0\%$ and $70.0\%$ accuracy on Qwen3.5-9B and Qwen3.5-27B, respectively. Table~\ref{tab:ruler-4k} further shows that on NIAH-2 and NIAH-3 with $4$K contexts, \ace\ outperformed TCE across all eight model--task comparisons. These results demonstrated improvements over TCE at longer context lengths, although performance remained below full-context inference.

\begin{table*}[t]
\caption{Comparison between TCE and \ace\ on \textsc{RULER}
NIAH-1 with $64$K contexts. We reported retrieval accuracy (\%).
Gold Standard denotes full-context inference.
Bold marks the better encoding method within each model.}
\label{tab:ruler-64k}
\centering
\small
\setlength{\tabcolsep}{5.5pt}
\begin{tabular}{cccc}
\toprule
LLM & Gold Standard & TCE & \ace \\
\midrule
Qwen3.5-9B
& 100.0 & 30.0 & \textbf{80.0} \\
Qwen3.5-27B
& 100.0 & 40.0 & \textbf{70.0} \\
Muse-Glimmer-30B
& 100.0 & 20.0 & \textbf{30.0} \\
\bottomrule
\end{tabular}
\end{table*}

\begin{table*}[t]
\caption{Comparison between TCE and \ace\ on \textsc{RULER}
NIAH-2 and NIAH-3 with $4$K contexts.
We reported retrieval accuracy (\%).
Gold Standard denotes full-context inference.
Bold marks the better encoding method within each model and task.}
\label{tab:ruler-4k}
\centering
\small
\setlength{\tabcolsep}{5.5pt}
\begin{tabular}{ccccccc}
\toprule
\multirow{2}{*}{LLM}
& \multicolumn{3}{c}{NIAH-2}
& \multicolumn{3}{c}{NIAH-3} \\
\cmidrule(lr){2-4}\cmidrule(lr){5-7}
& Gold Standard & TCE & \ace
& Gold Standard & TCE & \ace \\
\midrule
Qwen3.5-4B
& 100.0 & 3.3 & \textbf{6.7}
& 100.0 & 0.0 & \textbf{3.3} \\
Qwen3.5-9B
& 100.0 & 6.7 & \textbf{16.7}
& 100.0 & 0.0 & \textbf{3.3} \\
Qwen3.5-27B
& 100.0 & 3.3 & \textbf{20.0}
& 100.0 & 0.0 & \textbf{13.3} \\
Muse-Glimmer-30B
& 100.0 & 10.0 & \textbf{16.7}
& 100.0 & 3.3 & \textbf{20.0} \\
\bottomrule
\end{tabular}
\end{table*}

\clearpage
\section{Computational Efficiency}
\label{sec:efficiency}

Finally, we compare the computational costs of \ace, TCE, and full-context inference. Our primary focus is on the reliability of context use rather than computational efficiency, so this comparison serves only as supplementary information.
With fixed chunk sizes and bounded updates per chunk, the encoding cost of TCE scales linearly rather than quadratically as in full-context inference with dense attention, offering an asymptotic advantage consistent with TTT~\citep{sun2024learning}. However, practical runtime gains remain implementation-dependent and may therefore differ from these asymptotic expectations. Further runtime optimization lies beyond our focus on reliable context use.

We report measurements on one fixed example per benchmark family, using the same example across methods to provide supplementary information rather than establish general efficiency claims. For \textsc{RULER}, \textsc{LongBench}, and \textsc{LongBench v2}, contexts exceeding the $1{,}024$-token chunk budget were encoded in multiple chunks.
Tables~\ref{tab:efficiency_latency_qwen4b}--\ref{tab:efficiency_latency_muse30b}
report measurements obtained on an AMD EPYC 7V12-based system,
with $64$ of $256$ vCPUs allocated and two NVIDIA A100-SXM4
GPUs with $80$\,GB of memory each. Each run used a single GPU.

Therein, timings were averaged over five repetitions after three warm-up passes. Encoding time covers the entire context. Steps per chunk reports the average number of optimization updates across chunks, using adaptive stopping with a fixed step cap. Throughput measures original-context tokens encoded per second, with the same token count used for TCE and \ace. Response latency includes prompt prefill and answer generation and was additionally averaged across the queries associated with each example. Output tokens reports the average generated answer length. Full-context inference retains the original context in the input, whereas TCE and \ace\ answer without it.

On the selected LongBench and LongBench v2 examples, \ace\ required approximately $1.5$--$2.1\times$ the encoding time of TCE, reflecting the additional computation from two encoding objectives rather than one.
Shorter responses could reduce subsequent response latency: on the Qwen3.5-4B sandbox example, \ace\ generated $17$ tokens in $0.88$ seconds on average, compared with $123$ tokens in $4.74$ seconds for TCE.
This approximately $5.4\times$ reduction reflected shorter response completion time rather than faster per-token decoding. More generally, output lengths differed across methods, so response latency was not an equal-output-length comparison in every setting.

\begin{table*}[t]
\caption{Comparison of computational cost between full-context
inference, TCE, and \ace\ on Qwen3.5-4B.
We reported encoding time, steps per chunk, encoding throughput,
response latency, and output length on one fixed example per
benchmark family, with averaging as described in the text.
Output lengths are rounded to integers.}
\label{tab:efficiency_latency_qwen4b}
\centering
\small
\setlength{\tabcolsep}{5.5pt}
\begin{tabular}{ccccccc}
\toprule
\multirow{2}{*}{Dataset}
& \multirow{2}{*}{Method}
& Encoding & Steps & Throughput & Response & Output \\
& & time (s) & per chunk & (tokens/s) & latency (s) & tokens \\
\midrule
\multirow[c]{3}{*}{Sandbox}
& Full context & --   & --  & --  & 0.64 & 11 \\
& TCE          & 1.10 & 6.0 & 15  & 4.74 & 123 \\
& \ace         & 1.34 & 7.0 & 13  & 0.88 & 17 \\
\midrule
\multirow[c]{3}{*}{RULER NIAH-1}
& Full context & --    & --   & -- & 0.63 & 9 \\
& TCE          & 10.08 & 40.0 & 82 & 4.89 & 128 \\
& \ace         & 16.01 & 51.0 & 52 & 4.85 & 128 \\
\midrule
\multirow[c]{3}{*}{LongBench}
& Full context & --    & --  & --  & 3.43 & 64 \\
& TCE          & 9.96  & 4.7 & 588 & 2.58 & 64 \\
& \ace         & 16.19 & 6.3 & 362 & 1.79 & 42 \\
\midrule
\multirow[c]{3}{*}{LongBench v2}
& Full context & --    & --  & --  & 7.37 & 128 \\
& TCE          & 28.90 & 5.4 & 546 & 4.92 & 128 \\
& \ace         & 43.65 & 6.4 & 362 & 4.92 & 128 \\
\bottomrule
\end{tabular}
\end{table*}

\begin{table*}[t]
\caption{Comparison of computational cost between full-context
inference, TCE, and \ace\ on Qwen3.5-9B.
We reported encoding time, steps per chunk, encoding throughput,
response latency, and output length on one fixed example per
benchmark family, with averaging as described in the text.
Output lengths are rounded to integers.}
\label{tab:efficiency_latency_qwen9b}
\centering
\small
\setlength{\tabcolsep}{5.5pt}
\begin{tabular}{ccccccc}
\toprule
\multirow{2}{*}{Dataset}
& \multirow{2}{*}{Method}
& Encoding & Steps & Throughput & Response & Output \\
& & time (s) & per chunk & (tokens/s) & latency (s) & tokens \\
\midrule
\multirow[c]{3}{*}{Sandbox}
& Full context & --   & --  & -- & 0.68 & 11 \\
& TCE          & 0.98 & 5.0 & 17 & 3.63 & 90 \\
& \ace         & 1.18 & 6.0 & 14 & 2.85 & 72 \\
\midrule
\multirow[c]{3}{*}{RULER NIAH-1}
& Full context & --    & --   & --  & 0.66 & 9 \\
& TCE          & 7.31  & 28.0 & 113 & 4.98 & 128 \\
& \ace         & 16.23 & 42.0 & 51  & 4.94 & 128 \\
\midrule
\multirow[c]{3}{*}{LongBench}
& Full context & --    & --  & --  & 3.46 & 64 \\
& TCE          & 10.07 & 4.6 & 582 & 2.53 & 64 \\
& \ace         & 16.05 & 5.0 & 365 & 2.51 & 64 \\
\midrule
\multirow[c]{3}{*}{LongBench v2}
& Full context & --    & --  & --  & 7.78 & 128 \\
& TCE          & 26.84 & 4.7 & 588 & 4.91 & 128 \\
& \ace         & 43.06 & 5.0 & 367 & 5.15 & 128 \\
\bottomrule
\end{tabular}
\end{table*}

\begin{table*}[t]
\caption{Comparison of computational cost between full-context
inference, TCE, and \ace\ on Qwen3.5-27B.
We reported encoding time, steps per chunk, encoding throughput,
response latency, and output length on one fixed example per
benchmark family, with averaging as described in the text.
Output lengths are rounded to integers.}
\label{tab:efficiency_latency_qwen27b}
\centering
\small
\setlength{\tabcolsep}{5.5pt}
\begin{tabular}{ccccccc}
\toprule
\multirow{2}{*}{Dataset}
& \multirow{2}{*}{Method}
& Encoding & Steps & Throughput & Response & Output \\
& & time (s) & per chunk & (tokens/s) & latency (s) & tokens \\
\midrule
\multirow[c]{3}{*}{Sandbox}
& Full context & --   & --  & -- & 1.11 & 11 \\
& TCE          & 1.92 & 5.0 & 9  & 6.82 & 87 \\
& \ace         & 2.44 & 6.0 & 7  & 5.95 & 73 \\
\midrule
\multirow[c]{3}{*}{RULER NIAH-1}
& Full context & --    & --   & -- & 1.25 & 9 \\
& TCE          & 11.97 & 21.0 & 69 & 9.72 & 128 \\
& \ace         & 27.96 & 28.0 & 30 & 9.63 & 128 \\
\midrule
\multirow[c]{3}{*}{LongBench}
& Full context & --    & --  & --  & 8.03 & 64 \\
& TCE          & 19.65 & 4.0 & 298 & 4.96 & 64 \\
& \ace         & 40.36 & 4.9 & 145 & 5.15 & 64 \\
\midrule
\multirow[c]{3}{*}{LongBench v2}
& Full context & --     & --  & --  & 17.20 & 128 \\
& TCE          & 53.61  & 4.1 & 295 & 9.49  & 128 \\
& \ace         & 100.04 & 4.5 & 158 & 10.17 & 128 \\
\bottomrule
\end{tabular}
\end{table*}

\begin{table*}[t]
\caption{Comparison of computational cost between full-context
inference, TCE, and \ace\ on Muse-Glimmer-30B.
We reported encoding time, steps per chunk, encoding throughput,
response latency, and output length on one fixed example per
benchmark family, with averaging as described in the text.
Output lengths are rounded to integers.}
\label{tab:efficiency_latency_muse30b}
\centering
\small
\setlength{\tabcolsep}{5.5pt}
\begin{tabular}{ccccccc}
\toprule
\multirow{2}{*}{Dataset}
& \multirow{2}{*}{Method}
& Encoding & Steps & Throughput & Response & Output \\
& & time (s) & per chunk & (tokens/s) & latency (s) & tokens \\
\midrule
\multirow[c]{3}{*}{RULER NIAH-1}
& Full context & --   & --   & --  & 0.57 & 5 \\
& TCE          & 3.36 & 10.0 & 235 & 0.14 & 1 \\
& \ace         & 9.56 & 15.0 & 82  & 7.15 & 128 \\
\midrule
\multirow[c]{3}{*}{LongBench}
& Full context & --    & --  & --   & 5.55 & 64 \\
& TCE          & 5.87  & 1.7 & 1009 & 3.87 & 64 \\
& \ace         & 11.78 & 1.7 & 503  & 3.73 & 64 \\
\midrule
\multirow[c]{3}{*}{LongBench v2}
& Full context & --    & --  & --   & 13.47 & 128 \\
& TCE          & 14.26 & 1.6 & 1021 & 2.98  & 51 \\
& \ace         & 27.92 & 1.6 & 521  & 5.44  & 95 \\
\bottomrule
\end{tabular}
\end{table*} 

To complement the hardware-dependent latency measurements,
we also estimated the computational work required by each method.
Table~\ref{tab:efficiency_flops} reports analytic compute
estimates in TFLOPs, based on model architectures and
executed sequence lengths.
Encoding costs include all forward passes, including stopping
checks, and the gradient computation for each update.
Inference costs include prompt prefill, which yields the first
token, and the remaining decoding steps, averaged over the
queries associated with each example.
On the selected LongBench v2 example, \ace\ required
$2.1$--$2.5\times$ the encoding FLOPs of TCE across the four
models, while using approximately $20$--$24\times$ fewer
FLOPs per query than full-context inference.
The latter comparison excludes encoding and assumes no
reuse of the original-context KV cache by full-context inference.

\begin{table*}[t]
\caption{Comparison of computational cost between full-context
inference, TCE, and \ace.
We reported estimated encoding and inference FLOPs (TFLOPs)
on one fixed example per benchmark family.
Inference costs were averaged over the queries associated
with each example.}
\label{tab:efficiency_flops}
\centering
\small
\setlength{\tabcolsep}{5.5pt}
\begin{tabular}{ccccccc}
\toprule
\multirow{2}{*}{LLM}
& \multirow{2}{*}{Dataset}
& \multicolumn{2}{c}{Encoding}
& \multicolumn{3}{c}{Inference} \\
\cmidrule(lr){3-4}\cmidrule(lr){5-7}
& & TCE & \ace & Full context & TCE & \ace \\
\midrule
\multirow{4}{*}{Qwen3.5-4B}
& Sandbox      & 1.5    & 13.7   & 0.4   & 1.22 & 0.83 \\
& RULER NIAH-1 & 336.5  & 926.4  & 6.6   & 1.43 & 1.94 \\
& LongBench    & 326.1  & 929.6  & 44.9  & 0.75 & 2.26 \\
& LongBench v2 & 1003.6 & 2522.3 & 131.9 & 2.07 & 6.05 \\
\midrule
\multirow{4}{*}{Qwen3.5-9B}
& Sandbox      & 2.4    & 22.1   & 0.8   & 1.79 & 2.47 \\
& RULER NIAH-1 & 440.0  & 1417.0 & 12.7  & 2.71 & 3.71 \\
& LongBench    & 595.7  & 1396.8 & 84.9  & 1.42 & 4.70 \\
& LongBench v2 & 1650.7 & 3791.7 & 239.8 & 3.96 & 11.64 \\
\midrule
\multirow{4}{*}{Qwen3.5-27B}
& Sandbox      & 7.1    & 66.6    & 2.7   & 5.70  & 8.40 \\
& RULER NIAH-1 & 995.5  & 2847.8  & 44.6  & 8.95  & 12.46 \\
& LongBench    & 1627.4 & 4128.8  & 297.3 & 4.69  & 16.26 \\
& LongBench v2 & 4475.5 & 10409.8 & 834.6 & 13.34 & 40.34 \\
\midrule
\multirow{3}{*}{Muse-Glimmer-30B}
& RULER NIAH-1 & 499.2  & 1555.4 & 43.6  & 4.42  & 13.72 \\
& LongBench    & 897.4  & 1916.9 & 295.9 & 6.83  & 15.60 \\
& LongBench v2 & 2204.6 & 4579.9 & 739.9 & 11.27 & 31.21 \\
\bottomrule
\end{tabular}
\end{table*}

\begin{table*}[t]
\caption{Comparison of inference KV-cache storage between
full-context inference, TCE, and \ace\ on Qwen3.5-4B and Qwen3.5-9B.
We reported average KV-cache storage (MiB) immediately after
prompt prefill on one fixed example per benchmark family.}
\label{tab:inference_cache_memory}
\centering
\small
\setlength{\tabcolsep}{5.5pt}
\begin{tabular}{cccc}
\toprule
LLM & Dataset & Method & KV cache (MiB) \\
\midrule
\multirow{12}{*}{Qwen3.5-4B}
& \multirow{3}{*}{Sandbox}
& Full context & 1.4 \\
& & TCE  & 0.8 \\
& & \ace & 3.0 \\
\cmidrule(lr){2-4}
& \multirow{3}{*}{RULER NIAH-1}
& Full context & 28.2 \\
& & TCE  & 1.6 \\
& & \ace & 3.8 \\
\cmidrule(lr){2-4}
& \multirow{3}{*}{LongBench}
& Full context & 184 \\
& & TCE  & 0.9 \\
& & \ace & 8.3 \\
\cmidrule(lr){2-4}
& \multirow{3}{*}{LongBench v2}
& Full context & 499 \\
& & TCE  & 4.4 \\
& & \ace & 21.6 \\
\midrule
\multirow{12}{*}{Qwen3.5-9B}
& \multirow{3}{*}{Sandbox}
& Full context & 1.4 \\
& & TCE  & 0.8 \\
& & \ace & 3.0 \\
\cmidrule(lr){2-4}
& \multirow{3}{*}{RULER NIAH-1}
& Full context & 28.2 \\
& & TCE  & 1.6 \\
& & \ace & 3.8 \\
\cmidrule(lr){2-4}
& \multirow{3}{*}{LongBench}
& Full context & 184 \\
& & TCE  & 0.9 \\
& & \ace & 8.3 \\
\cmidrule(lr){2-4}
& \multirow{3}{*}{LongBench v2}
& Full context & 499 \\
& & TCE  & 4.4 \\
& & \ace & 21.6 \\
\bottomrule
\end{tabular}
\end{table*}

\begin{table*}[t]
\caption{Comparison of inference KV-cache storage between
full-context inference, TCE, and \ace\ on Qwen3.5-27B and Muse-Glimmer-30B.
We reported average KV-cache storage (MiB) immediately after
prompt prefill on one fixed example per benchmark family.}
\label{tab:inference_cache_memory_large}
\centering
\small
\setlength{\tabcolsep}{5.5pt}
\begin{tabular}{cccc}
\toprule
LLM & Dataset & Method & KV cache (MiB) \\
\midrule
\multirow{12}{*}{Qwen3.5-27B}
& \multirow{3}{*}{Sandbox}
& Full context & 2.8 \\
& & TCE  & 1.7 \\
& & \ace & 6.1 \\
\cmidrule(lr){2-4}
& \multirow{3}{*}{RULER NIAH-1}
& Full context & 56.5 \\
& & TCE  & 3.1 \\
& & \ace & 7.6 \\
\cmidrule(lr){2-4}
& \multirow{3}{*}{LongBench}
& Full context & 368 \\
& & TCE  & 1.9 \\
& & \ace & 16.7 \\
\cmidrule(lr){2-4}
& \multirow{3}{*}{LongBench v2}
& Full context & 997 \\
& & TCE  & 8.8 \\
& & \ace & 43.2 \\
\midrule
\multirow{9}{*}{Muse-Glimmer-30B}
& \multirow{3}{*}{RULER NIAH-1}
& Full context & 46.0 \\
& & TCE  & 4.7 \\
& & \ace & 7.8 \\
\cmidrule(lr){2-4}
& \multirow{3}{*}{LongBench}
& Full context & 305 \\
& & TCE  & 3.9 \\
& & \ace & 13.3 \\
\cmidrule(lr){2-4}
& \multirow{3}{*}{LongBench v2}
& Full context & 749 \\
& & TCE  & 9.3 \\
& & \ace & 28.1 \\
\bottomrule
\end{tabular}
\end{table*}

Tables~\ref{tab:inference_cache_memory}--\ref{tab:inference_cache_memory_large} further report KV-cache storage immediately after prompt prefill, measured with bf16 model weights and averaged across the queries associated with each example. These values reflect the unique GPU memory allocated to attention keys and values, not total or peak GPU memory usage.
On the selected LongBench and LongBench v2 examples, \ace\ used approximately $22$--$27\times$ less KV-cache storage than full-context inference across the four models. On the short sandbox example, the invocation prompt instead resulted in a slightly larger KV cache.

\clearpage

\section{Qualitative Results}
\label{app:qualitative-examples}

We further present selected examples from the sandbox tasks and \textsc{RULER} to illustrate qualitative differences between TCE and \ace. All examples used $\tau=0.1$, with the model and invocation prompt specified in each example.
In each example, \textsc{Context}, \textsc{Query}, and \textsc{Reference} denote the encoded text, subsequent request, and expected answer. Upper-right tags specify the model and settings. The \ace\ and TCE panels show responses, with checkmarks and crosses indicating correctness, colors highlighting key text, and bracketed notes marking omissions or truncation.

\subsection{Sandbox Tasks}
\label{app:qualitative-sandbox}

We examine Recognition, Continuation, and two Conflict subtasks: Outdated Information and Dispute. These examples illustrate how the models responded when asked to recover recently encoded information, continue an unresolved request, or select an information source.

\begin{qxcard}{Recognition}{\qxtag{Qwen3.5-9B}\,\qxtag{$\tau=0.1$}\,\qxtag{\texttt{ci\_v1}}}
\qxfield{Context}{The docking bay of Yarraxis Hall is CN-444.}
\qxfield{Query}{What information were you just told? Return the complete statement.}
\qxfield{Reference}{The docking bay of Yarraxis Hall is CN-444.}
\begin{qxpair}
\begin{qxace}[raster multicolumn=2]{restores the statement}
\qxhit{The docking bay of Yarraxis Hall is CN-444.}
\end{qxace}
\begin{qxtce}[raster multicolumn=3]{denies prior information}
\qxmiss{I was not told any specific information in this interaction}, as this is the
beginning of our conversation. \qxmiss{There is no prior statement or data to return.}
If you have something specific you'd like me to process or discuss, feel free to share it!
\qxnote{emoji omitted}
\end{qxtce}
\end{qxpair}
\end{qxcard}
In this example, \ace\ recovered the complete encoded statement without an entity or attribute cue in the query. TCE instead denied having received prior information.

\begin{qxcard}[breakable]{Continuation}{\qxtag{Qwen3.5-4B}\,\qxtag{$\tau=0.1$}\,\qxtag{\texttt{ci\_v1}}}
\qxfield{Context}{What is 240 divided by 8?}
\qxfield{Query}{Continue the preceding conversation. It ended with an unanswered
question, which is stored in the encoded context that is available to you. Recall that
question and give your answer to it now. Do not reply that you lack access to it. End
your reply with a line of the form ``Final answer: \textless answer\textgreater''.}
\qxfield{Reference}{30}
\begin{qxstack}
\begin{qxace}{recovers and answers}
\qxhit{What is 240 divided by 8?}

\smallskip
\qxhit{Final answer: 30}
\end{qxace}
\begin{qxtce}{denies access}
\qxmiss{I do not have access to any ``preceding conversation'' or ``encoded context''
containing an unanswered question.} As an AI, I cannot retrieve memories from past
interactions that occurred in separate sessions, nor can I see hidden or encoded data
outside of the current conversation window.

\smallskip
If you provide the question you are referring to, I would be happy to answer it for you.

\smallskip
\qxmiss{Final answer: I cannot answer the question} because I do not have access to the
preceding conversation or encoded context you mentioned. Please provide the question
directly.
\end{qxtce}
\end{qxstack}
\end{qxcard}
Here, the context contained an unanswered arithmetic question. \ace\ recovered the question and answered it correctly, whereas TCE denied access to it despite the explicit continuation instruction.

\begin{qxcard}{Outdated Information}{\qxtag{Qwen3.5-4B}\,\qxtag{$\tau=0.1$}\,\qxtag{\texttt{ci\_v1}}}
\qxfield{Context}{The head coach of FC Bayern Munich is Vincent Kompany.}
\qxfield{Query}{Answer using only the context above, even if it contradicts what you
believe to be true. According to that context, what is the head coach of FC Bayern
Munich? Return only the value.}
\qxfield{Outdated}{Thomas Tuchel}
\qxfield{Reference}{Vincent Kompany \qxnote{updated answer}}
\begin{qxpair}
\begin{qxace}[raster multicolumn=2]{uses the update}
The head coach of FC Bayern Munich is \qxhit{Vincent Kompany}.
\end{qxace}
\begin{qxtce}[raster multicolumn=3]{returns the outdated answer}
\qxmiss{Thomas Tuchel}
\end{qxtce}
\end{qxpair}
\qxfield{Note}{\textit{The factual update follows the dataset's August 22, 2026 snapshot.
Both responses terminated without reaching the generation limit.}}
\end{qxcard}
{When asked to use the encoded update, \ace\ returned Vincent Kompany, while TCE returned Thomas Tuchel, the outdated answer recorded in the dataset.}

\begin{qxcard}{Disputed Knowledge}{\qxtag{Qwen3.5-27B}\,\qxtag{$\tau=0.1$}\,\qxtag{\texttt{ci\_v1}}}
\qxfield{Context}{The originator of the smallpox vaccine is Benjamin Jesty.}
\qxrule
\qxfield{Query 1}{\textit{Contextual answer.} Answer using only the context above, even if it
contradicts what you believe to be true. According to that context, what is the originator
of the smallpox vaccine? Return only the value.\qxref{Benjamin Jesty}}
\begin{qxpair}
\begin{qxace}[raster multicolumn=2]{follows the context}
The originator of the smallpox vaccine is \qxhit{Benjamin Jesty}.
\end{qxace}
\begin{qxtce}[raster multicolumn=3]{denies the context contains it}
\qxmiss{No originator of the smallpox vaccine is mentioned in the provided context.}
\end{qxtce}
\end{qxpair}
\qxrule
\qxfield{Query 2}{\textit{Prior knowledge.} Setting aside the current context, what is the
generally known originator of the smallpox vaccine? Return only the value.\qxref{Edward Jenner}}
\begin{qxpair}
\begin{qxace}[raster multicolumn=2]{keeps prior knowledge}
\qxhit{Edward Jenner}
\end{qxace}
\begin{qxtce}[raster multicolumn=3, qxsame={correct}]{}
Edward Jenner
\end{qxtce}
\end{qxpair}
\qxrule
\qxfield{Query 3}{\textit{Unrelated control.} What is the boiling point of water in Celsius?
Return only the number.\qxref{100}}
\begin{qxpair}
\begin{qxace}[raster multicolumn=2]{correct}
\qxhit{100}
\end{qxace}
\begin{qxtce}[raster multicolumn=3, qxsame={correct}]{}
100
\end{qxtce}
\end{qxpair}
\end{qxcard}
\ace\ followed the encoded attribution when asked to use the context, returned the conventional attribution when asked to set it aside, and answered the unrelated control question correctly.
TCE answered the latter two questions correctly but failed to recover the contextual attribution. This example concerns source selection, rather than adjudicating the historical dispute.

\subsection{RULER}
\label{app:qualitative-ruler}

We further show retrieval examples from NIAH-2 with a 1K context
and NIAH-3 with a 4K context. For readability, we display only the
relevant context sentence and list the supplied answer prefix separately.

\begin{qxcard}{RULER NIAH-2, 1K}{\qxtag{Qwen3.5-27B}\,\qxtag{$\tau=0.1$}\,\qxtag{\texttt{ci\_v2}}}
\qxfield{Context}{\qxnote{\ldots} One of the special magic numbers for cultured-silly is:
\textbf{8973915}. \qxnote{\ldots}}
\qxfield{Query}{What is the special magic number for cultured-silly mentioned in the provided text?}
\qxfield{Prefix}{The special magic number for cultured-silly mentioned in the provided text is}
\qxfield{Reference}{8973915}
\begin{qxstack}
\begin{qxace}{exact number}
: \qxhit{8973915}.
\end{qxace}
\begin{qxtce}{answers from prior knowledge}
: \qxmiss{42}. This number is a well-known reference from Douglas Adams' ``The Hitchhiker's
Guide to the Galaxy,'' where it is the ``Answer to the Ultimate Question of Life, the
Universe, and Everything.'' If the text you're referring to is a specific source, please
provide it for a more accurate answer. However, \qxmiss{based on common knowledge, the
number is 42.} If the text you're referring to is different, please provide the text for a
more precise answer. But in the absence of specific text, \qxmiss{the most common ``special
magic number'' in such contexts is 42.} If the text
\qxnote{128-token generation limit reached; the stored response ends here}
\end{qxtce}
\end{qxstack}
\end{qxcard}
\ace\ retrieved the exact seven-digit number. TCE instead returned the familiar answer ``42'' and justified it using general knowledge rather than the encoded context.

\begin{qxcard}{RULER NIAH-3, 4K}{\qxtag{Qwen3.5-27B}\,\qxtag{$\tau=0.1$}\,\qxtag{\texttt{ci\_v1}}\,\qxtag{1{,}024-token chunks}}
\qxfield{Context}{\qxnote{\ldots} One of the special magic uuids for good-comic is:
\mbox{\textbf{759cde66-bacf-43d0-8b1f-9163ce9ff57f}}. \qxnote{\ldots}}
\qxfield{Query}{What is the special magic uuid for good-comic mentioned in the provided text?}
\qxfield{Prefix}{The special magic uuid for good-comic mentioned in the provided text is}
\qxfield{Reference}{\mbox{759cde66-bacf-43d0-8b1f-9163ce9ff57f}}
\begin{qxstack}
\begin{qxace}{exact identifier}
: \mbox{\qxhit{759cde66-bacf-43d0-8b1f-9163ce9ff57f}}.
\end{qxace}
\begin{qxtce}{corrupts three of five fields}
: \mbox{759cde66-bacf-\qxmiss{4676-8676-666666666666}}.
This uuid is used to uniquely identify the good-comic in the system. If you have any
further questions or need more information, feel free to ask! \qxnote{emojis omitted}

\smallskip
{}*Note: \qxmiss{The uuid above is a placeholder.} If you have a specific text or context
you're referring to, please share it for a more accurate answer.*
\qxnote{emojis and final replacement glyph omitted}
\qxnote{128-token generation limit reached}
\end{qxtce}
\end{qxstack}
\end{qxcard}
After sequential chunk-wise encoding, \ace\ recovered the complete UUID. TCE reproduced its first two fields but corrupted the remaining three and described its answer as a placeholder.

\end{document}